\documentclass[runningheads]{llncs}
\usepackage{graphicx}
\usepackage{tikz}
\usepackage{comment}
\usepackage{amsmath,amssymb} % define this before the line numbering.
\usepackage{color}
\usepackage{hyperref}

\usepackage[accsupp]{axessibility}  % Improves PDF readability for those with disabilities.

\usepackage[width=122mm,left=12mm,paperwidth=146mm,height=193mm,top=12mm,paperheight=217mm]{geometry}

\usepackage{subfiles}
\usepackage{caption}
\usepackage{subcaption}

\usepackage[sort,nocompress]{cite} % nocompress disables numeber ranges,e.g. [9-11, ..., 94-96]

\usepackage{enumitem}

\usepackage{xspace}
\newcommand*{\datasetfullname}{Caltech Fish Counting Dataset\@\xspace}
\newcommand*{\datasetname}{CFC\@\xspace}

\usepackage{booktabs, multirow} % for borders and merged ranges
\usepackage{soul}% for underlines
\usepackage{makecell}
\newcommand{\mytilde}{\raise.17ex\hbox{$\scriptstyle\mathtt{\sim}$}}

\newcommand{\samelineand}{\enspace\enspace\stepcounter{@inst}}

\newcommand{\newsamelineand}{\qquad\stepcounter{@inst}%
 $^{\the@inst}$\enspace\ignorespaces}%

\begin{document}
% \renewcommand\thelinenumber{\color[rgb]{0.2,0.5,0.8}\normalfont\sffamily\scriptsize\arabic{linenumber}\color[rgb]{0,0,0}}
% \renewcommand\makeLineNumber {\hss\thelinenumber\ \hspace{6mm} \rlap{\hskip\textwidth\ \hspace{6.5mm}\thelinenumber}}
% \linenumbers
\pagestyle{headings}
\mainmatter
\def\ECCVSubNumber{5272}  % Insert your submission number here

% \title{The \datasetfullname: A Multi-Object Tracking and Counting Benchmark}
\title{The \datasetfullname: A Benchmark for Multiple-Object Tracking and Counting}

% INITIAL SUBMISSION 
\begin{comment}
\titlerunning{ECCV-22 submission ID \ECCVSubNumber} 
\authorrunning{ECCV-22 submission ID \ECCVSubNumber} 
\author{Anonymous ECCV submission}
\institute{Paper ID \ECCVSubNumber}
\end{comment}
%******************

% CAMERA READY SUBMISSION
%\begin{comment}
\titlerunning{The \datasetfullname}
% If the paper title is too long for the running head, you can set
% an abbreviated paper title here
%
\author{
Justin Kay\inst{1,5} \and
Peter Kulits\inst{1} \and
Suzanne Stathatos\inst{1} \and
Siqi Deng\inst{2} \and
Erik Young\inst{3} \and
Sara Beery\inst{1} \and
Grant Van Horn\inst{4}\index{Van Horn, Grant} \and
Pietro Perona\inst{1,2}
}
\authorrunning{J. Kay et al.}
% First names are abbreviated in the running head.
% If there are more than two authors, 'et al.' is used.
%
\institute{
California Institute of Technology \samelineand
\inst{2}\enspace AWS AI Labs \samelineand
\inst{3}\enspace Trout Unlimited \and
Cornell University \samelineand
\inst{5}\enspace Ai.Fish
}

%\end{comment}
%******************
\maketitle
\begin{abstract}
%%%% PREVIOUS VERSION %%%%
% We present the \datasetfullname (\datasetname), a large-scale dataset for detecting, tracking, and counting fish in sonar videos. 
% We identify sonar videos as a rich source of data for tackling domain generalization in the context of multiple-object tracking (MOT) and counting.
% Current large-scale datasets for MOT and video-based counting are largely restricted to people in urban environments. 
% While animal-centric datasets do exist, they are small and do not support both tracking and counting.
% Furthermore, no large-scale MOT or counting dataset is designed to test the generalization ability of algorithms to unseen evaluation settings.
% We specifically designed \datasetname with this generalization task in mind.
% With over half a million annotations in over 1,500 videos, \datasetname allows researchers to train MOT algorithms and evaluate them on unseen test scenarios. 
% We perform extensive baseline experiments and identify key challenges and opportunities for advancing the state of the art in the generalization of MOT and counting.
We present the \datasetfullname (\datasetname), a large-scale dataset for detecting, tracking, and counting fish in sonar videos. We identify sonar videos as a rich source of data for advancing low signal-to-noise computer vision applications and tackling domain generalization in multiple-object tracking (MOT) and counting. In comparison to existing MOT and counting datasets, which are largely restricted to videos of people and vehicles in cities, CFC is sourced from a natural-world domain where targets are not easily resolvable and appearance features cannot be easily leveraged for target re-identification. 
With over half a million annotations in over 1,500 videos sourced from seven different sonar cameras, CFC allows researchers to train MOT and counting algorithms and evaluate generalization performance at unseen test locations. We perform extensive baseline experiments and identify key challenges and opportunities for advancing the state of the art in generalization in MOT and counting.
%The dataset also introduces the computer vision community to a high-impact application in environmental conservation with real-world domain generalization challenges. 

\end{abstract}

%\pietro{Ideas for naming the dataset: \\
%(CAC-S) Caltech Animal Counting - Salmon\\
%(CAM-S) Caltech Animal Migration - Salmon\\
%(CAFCo) Caltech Anadromous Fish Counting}\\
%(CSC 2022) Caltech Salmon Counting\\
%(CFC 2022) Caltech Fish Counting\\
%(CaFi) Caltech Fish

\section{Introduction}

\noindent
Diverse and high-quality datasets collected from the natural world have enabled progress on fundamental computer vision tasks such as fine-grained visual categorization \cite{van2018inaturalist,wah2011caltech,nilsback2006visual,kumar2012leafsnap,van2015building,swanson2015snapshot,beery2019iwildcam,beery2018recognition,berg2014birdsnap} and individual re-identification \cite{parham2017animal,holmberg2009estimating,li2019amur}. This progress has had valuable impact in the same natural-world domains, and automated visual systems are now used in the field every day by ecologists, citizen scientists, and conservationists to improve the accuracy and efficiency of biodiversity monitoring efforts around the globe \cite{tuia2022perspectives,van2015building,kulits2021elephantbook,berger2017wildbook,ahumada2020wildlife,fennell2022use}. However, the range of computer vision tasks that these systems can perform is still limited, with most current algorithms and their supporting datasets focusing largely on visual classification in relatively high-quality imagery. Methods developed for existing datasets can fail to transfer to tasks where video analysis or non-RGB imagery is involved.

We present the \datasetfullname (\datasetname), a large video dataset containing over half a million annotations for detecting, tracking, and counting migrating fish in sonar video. In addition to providing a challenging benchmark in a novel application domain, the dataset allows for detailed study in three areas that have received limited attention from the computer vision community and lack supporting benchmarks:

%\begin{enumerate}[listparindent=1.5em]
    %\item 
    \textbf{1. Multiple-object tracking (MOT) in natural environments with animal targets.} Most existing MOT datasets focus on human~\cite{leal2015motchallenge,milan2016mot16,dendorfer2020mot20} and vehicle~\cite{wen2020ua,geiger2012we,sun2020scalability,yu2018bdd100k} tracking in cities. Animal targets provide a rich source of variability for developing trackers that are not biased toward urban domains, in addition to having numerous beneficial applications in conservation~\cite{tuia2022perspectives}, neuroscience~\cite{mathis2020deep}, and animal husbandry~\cite{fernandes2020image}. Furthermore, many state-of-the-art methods make extensive use of visual re-identification for performing association~\cite{zhang2020fairmot,revaud2016deepmatching,wojke2017simple}. In contrast, \datasetname provides a benchmark sourced from low signal-to-noise recording equipment in challenging natural-world environments where tracking targets are difficult to resolve from background clutter and each other, making frame-to-frame visual association less effective.
    
    The dataset is large, well-annotated, and challenging. It consists of 1,567 video sequences sourced from seven different sonar cameras on three rivers located in the U.S. states of Alaska and Washington. The videos are single-channel (i.e. grayscale), vary in resolution from 288x624 to 1,086x2,125, have frame rates between 6.7 and 13.3 frames per second, and are an average of 336 frames  (38 seconds) in duration. Tracking annotations were collected through a paid annotation service for 8,254 fish across 527k frames, totaling 516k bounding boxes in 16.7 hours of video.
    
    %\item 
    \textbf{2. Video-based counting.} Existing video counting benchmarks focus on crowd counting in urban environments \cite{wen2021detection,fang2019locality,chan2008privacy,change2013semi,dronecrowd_cvpr2021,fang2019locality} and emphasize density estimation over trajectory-based counting. Methods developed for these datasets have limitations in applications where information about individuals, such as size or direction of motion, is required, since crowd density is largely treated as a regional feature~\cite{sam2020locate}. The community is in need of a benchmark which can support both tracking and counting concurrently. \datasetname provides both a challenging MOT benchmark and an evaluation protocol for video-based counting that is motivated by a real-world metric. Ground-truth detections, trajectories, and counts are provided for every video sequence.
    
    %\item 
    \textbf{3. Generalization of tracking and counting methods to new domains.} While generalization in computer vision has been extensively studied in object recognition~\cite{csurka2017domain, Zhou2021DomainGI,koh2021wilds,beery2018recognition}, it is still relatively understudied within MOT and counting. \datasetname presents significant generalization challenges that are highlighted by the dataset design, providing a strong benchmark for the study of generalization and efficient adaptation in the context of MOT and counting. We enable this study by constraining training data to a single camera location, while sourcing test data from a variety of different out-of-sample rivers and cameras.
    
%\end{enumerate}

Finally, \datasetname is the first annotated video dataset sourced from the domain of fish counting in sonar, an application area with significant impacts in conservation ecology. 
Salmon are keystone species that support at least 137 other animal species and provide food and nutrients to a wide range of ecosystems during their seasonal migration \cite{rahrwhyprotectsalmon}. 
Sonar imaging provides a non-invasive way to monitor \textit{escapement}---the number of salmon returning home each season to spawn---helping inform sustainable fisheries management. Automation using computer vision could enable current sonar-based monitoring programs to scale from a few locations to entire watersheds. We hope that our dataset will encourage computer vision researchers to work on this high-impact challenge.

% could significantly reduce the human effort required, enabling current programs to scale from sparse points to entire watersheds. 
% Unfortunately, many salmon populations are threatened due to dams, hatcheries, loss of habitat, excessive fishing, and climate change. 
% Proper fisheries management can help restore and maintain healthy salmon populations by closely monitoring \textit{escapement}, the number of salmon returning home each season to spawn. 
% Imaging sonar provides a non-invasive way to collect escapement data, but the labor requirements of manual video analysis are a key barrier to scale. 
% Automation using computer vision could significantly reduce the human effort required, enabling current programs to scale from sparse points to entire watersheds. 
% We hope that our dataset will encourage computer vision researchers to work on this high-impact challenge.

Our contributions are: (1) a large and challenging dataset for tracking and counting in video that enables the study of generalizing algorithms to new locations; 
% (2) an evaluation protocol that benchmarks algorithms against a real-world metric; 
(2) an evaluation protocol 
%that benchmarks algorithms against a counting protocol developed by field biologists.
that mimics the procedure used by field technicians when manually counting fish in sonar video;
%in the target application; 
% an operational counting metric used in the target application;
%a real-world counting metric used in the target application;
% (3) a strong baseline algorithm that utilizes high-performing detection and tracking components along with a novel input structure.
% (3) a strong baseline algorithm for tracking and counting that utilizes a novel input structure to improve generalization performance.
(3) a baseline method that utilizes a novel input structure to improve generalization performance at unseen test locations. 
The dataset and evaluation code are available \href{https://github.com/visipedia/caltech-fish-counting}{here}.
% and evaluation code can be accessed  \href{https://github.com/visipedia/caltech-fish-counting}{here}.

\section{Related Work}

\label{sec-related}

\setlength{\tabcolsep}{4pt}
\begin{table}[t]
\begin{center}
\caption{\textbf{Comparison of video tracking and counting datasets.} \datasetname is the first dataset that supports all three tasks of interest: detection, tracking, and counting, with more tracking annotations than existing animal-centric tracking datasets. (* indicates annotations are points, not bounding boxes)
}
\label{table:datasets}
\begin{tabular}{l|r|r|r|l|c|c|c}
\hline\noalign{\smallskip}
Dataset & \multicolumn{1}{l|}{Vids} & \multicolumn{1}{l|}{Frames} & \multicolumn{1}{l|}{Annos} & \multicolumn{1}{l|}{Animals} & \multicolumn{1}{l|}{Detect} & \multicolumn{1}{l|}{Track} & \multicolumn{1}{l}{Count} \\
\noalign{\smallskip}
\hline\hline
\noalign{\smallskip}
% detect / track - alphabetical
 3D-ZeF \cite{pedersen20203d} & 8 & 28,800 & 86,400 & \checkmark (100\%) & \checkmark & \checkmark & \\ % not in original submission
BIRDSAI \cite{bondi2020birdsai} & 48 & 62,400 & 154,000 & \checkmark (78\%) & \checkmark & \checkmark & \\ %120k are animals
GMOT-40 \cite{bai2021gmot} & 40 & 9,643 & 256,341 &  \checkmark (38\%) & &  \checkmark & \\
MOT16 \cite{milan2016mot16} &14 &11,235 &292,733 & &\checkmark &\checkmark & \\
MOT20 \cite{dendorfer2020mot20} &8 &13,410 &1.65M & &\checkmark &\checkmark & \\
UA-DETRAC \cite{wen2020ua} &100 &140,000 &1.21M & &\checkmark &\checkmark & \\
TAO \cite{dave2020tao} & 2,907 & 4.44M & 332,401 & \checkmark (\mytilde10\%) & \checkmark & \checkmark &  \\

% count - alphabetical
AnimalDrone \cite{zhu2021graph} & 162 &53,644 & 4.05M* &\checkmark (100\%) & & &\checkmark  \\
DroneCrowd \cite{dronecrowd_cvpr2021} & 112 & 33,600 & 4.86M* & & & \checkmark &\checkmark \\
Crossing-line \cite{zhao2016crossing} &5 &3,100 &5,900* & & &\checkmark &\checkmark  \\
FDST \cite{fang2019locality} & 100 & 15,000 & 394,081* & & & &\checkmark  \\
Iowa DOT \cite{naphade20215th} &200 &90,000 &0 & & & &\checkmark \\
Mall \cite{chen2012feature} &1 &2,000 &62,315* & & & &\checkmark \\
UCSD \cite{chan2008privacy} &1 &2,000 &49,885* & & & &\checkmark \\
WorldExpo \cite{zhang2016data} &1,132 &3,980 &199,923* & & & & \checkmark \\
\midrule
\datasetname & 1,567 & 527,215 & 515,933 &\checkmark (100\%) &\checkmark &\checkmark &\checkmark \\
\hline
\end{tabular}
\end{center}
\end{table}

%\subsubsection{Multiple-Object Tracking (MOT)} 
\noindent
{\bf Multiple-Object Tracking (MOT).} Several popular benchmarks have supported recent progress in MOT, particularly in the domains of pedestrian and vehicle tracking \cite{leal2015motchallenge,milan2016mot16,dendorfer2020mot20,wen2020ua,geiger2012we,yu2018bdd100k,sun2020scalability}. Large-scale benchmarks for tracking animals in the wild are less common. TAO~\cite{dave2020tao} and GMOT-20~\cite{bai2021gmot} focus on tracking all foreground objects with limited class information, and include some animal tracking sequences. The BIRDSAI~\cite{bondi2020birdsai} dataset focuses on human and animal tracking in the wild, and---similar to ours---contains non-RGB (in their case, thermal infrared) sequences, though it is sourced from moving aerial cameras rather than in-situ monitoring devices. With over half a million MOT annotations, the \datasetname dataset is larger than existing benchmarks for animal tracking in the wild, with a unique additional focus on video-based counting and generalization challenges. See Tab.~\ref{table:datasets} for a comparison with prior work. 

Recent advancements in object detection~\cite{huang2017speed,zou2019object} have helped popularize the \textit{tracking-by-detection} paradigm in MOT, which divides the tracking problem into two steps: (1) an object detector predicts object locations (e.g. bounding boxes) in each frame, and (2) a tracker associates detections over time into object trajectories. While there has been significant progress in recent years (see recent surveys~\cite{ciaparrone2020deep,luo2021multiple}), there are two notable shortcomings. First, the tracking-by-detection paradigm implicitly assumes that detection is possible and accurate in each frame. This is not universally valid, and our dataset can stimulate research in algorithms that do not rely on this assumption due to the difficulty in resolving fish locations frame-to-frame. Second, much recent progress can be attributed to the development of complex visual-feature representations for target re-identification. These techniques are often domain-specific according to the benchmark dataset. For example, \cite{zheng2019joint} use a generative model to create synthetic pedestrian data consisting of various combinations of person appearance and structure information to achieve state-of-the-art performance, a technique that has been adopted by other top-performing MOT methods~\cite{hornakova2020lifted}. 
%Recently, motion-based methods have shown surprising efficacy without the use of appearance information for association, achieving state-of-the-art performance on top MOT benchmarks~\cite{zhang2021bytetrack}. 
Our dataset introduces a challenging MOT benchmark in which individuals are visually indistinct, offering little opportunity to make use of visual features for individual re-identification, which can motivate the development of tracking methods that are not dependent upon complex appearance matching.

%\subsubsection{Counting in Video}
\noindent
{\bf Counting in Video.} Datasets for object counting are predominantly image-based \cite{zhang2016single,zhang2016data,zhang2015cross,idrees2018composition,hsieh2017drone,wang2020nwpu,arteta2016counting,wu2014thermal,gemert2014nature,rey2017detecting,kellenberger2018detecting,shao2020cattle,weinstein2021remote,https://doi.org/10.3334/ornldaac/1832,beery2021iwildcam,norouzzadeh2018automatically,jones2018time}. Video-based counting datasets are more limited. Most are focused on estimating the number of people~\cite{wen2021detection,fang2019locality,chan2008privacy,change2013semi,dronecrowd_cvpr2021,fang2019locality} or animals~\cite{zhu2021graph} in crowded scenes, combining the challenges of crowd-density estimation and camera motion compensation. While a limited number of video counting datasets incorporate trajectory information \cite{dronecrowd_cvpr2021,zhao2016crossing}, existing benchmarks primarily model object locations as points and do not contain bounding box labels. \datasetname supports the study of all three tasks (detection, tracking, and counting), while containing over five times the number of annotated video frames as existing video counting benchmarks (see Tab.~\ref{table:datasets}).

Methods for video-based counting can be roughly divided into \textit{regression-based} methods, \textit{density-based} methods, and \textit{detection-based} methods. Regression-based methods attempt to predict counts directly by mapping image features to counting numbers \cite{moranduzzo2013automatic,kamenetsky2015aerial}, while density-based methods predict per-pixel crowd density in each frame and then analyze densities over time to obtain counts \cite{arteta2016counting,zhang2016single,zhao2016crossing,onoro2016towards,zhang2017fcn}. These methods are typically designed for counting large numbers of densely clustered objects where individual object detection is challenging. In contrast, detection-based methods utilize object detection in each frame to localize objects of interest and count them over time. These approaches typically employ a tracking-by-detection pipeline followed by counting based on either \textit{region-of-interest} (ROI)~\cite{mandal2020object,bui2020vehicle} or \textit{line-of-interest} (LOI)~\cite{kocamaz2016vision,ma2013crossing,zhao2016crossing}. ROI-based methods attempt to estimate the number of objects passing through a subregion of the frame, such as a traffic lane or onramp in the case of vehicle counting~\cite{bui2020vehicle}. LOI-based methods instead draw a virtual line through the field of view, counting objects when their trajectories intersect this line. In this work we use LOI-based counting, as it matches the approach currently used by fishery managers for counting fish in sonar~\cite{kenaiplan}.

%\subsubsection{Animal-Centric Datasets}
\label{sec-animal} 
\noindent
{\bf Animal-Centric Datasets.} Existing computer vision datasets in animal ecology primarily target the tasks of species classification~\cite{van2018inaturalist,wah2011caltech,nilsback2006visual,van2015building,swanson2015snapshot,beery2019iwildcam,berg2014birdsnap}, detection~\cite{beery2018recognition,beery2019iwildcam,swanson2015snapshot,jones2018time,tncchannel,pardo2021snapshot,yousif2019dynamic,zhang2016animal,anton2018monitoring,tabak2019machine,boom2014research,saleh2020realistic,cutter2015automated,ditria2021annotated,ds11,ds1,ds3}, and individual identification~\cite{parham2017animal,holmberg2009estimating,li2019amur}. These datasets consist predominantly of RGB imagery where visual features are key signals for recognition; in contrast, our dataset consists of single-channel sonar video, in which the animals of interest are difficult to distinguish from background, debris, and each other.
%\gvh{The previous sentence is great, we want more sentences like that (describing how our dataset fills a necessary gap missed by existing datasets)}
%Of these, the most closely related to our work are those sourced from in-situ devices such as camera traps \cite{beery2018recognition,beery2019iwildcam,swanson2015snapshot,jones2018time,tncchannel,pardo2021snapshot,yousif2019dynamic,zhang2016animal,anton2018monitoring,tabak2019machine} and remote underwater video (RUV) cameras \cite{boom2014research,saleh2020realistic,cutter2015automated,ditria2021annotated,ds11,ds1,ds3}, which depict the challenging conditions of the environments in which they are deployed: they may be low resolution due to hardware limitations, are taken from a range of environmental conditions, and do not necessarily provide a clear view of the species of interest. 

For this reason \datasetname shares some characteristics with video datasets for studying animal behavior, including mice \cite{austin2004knockout,bogue2020mouse,arac2019deepbehavior,hong2015automated,geuther2021action,geuther2019robust,sun2021multi}, rats \cite{van2013automated}, flies \cite{eyjolfsdottir2014detecting,eyjolfsdottir2016learning}, bees \cite{marstaller2019deepbees,boenisch2018tracking,rodriguez2018recognition,bozek2018towards}, and fish \cite{zhou2019fish,pedersen20203d,bruslund2020re}. As in our dataset, the visual similarities between individuals in these datasets mean that tracking must rely heavily on prediction of motion or behavior rather than using visual features for association. Our work, however, entails additional challenges not typically present in laboratory tracking and behavior study, such as complex background, difficult frame-by-frame detection, and unknown numbers of individuals in a scene.
%\gvh{again, great sentence}

%\subsubsection{Generalization in Computer Vision} 
\noindent
{\bf Generalization in Computer Vision.} \textit{Domain generalization} is a type of domain shift---i.e. a difference in training and test data distributions---in which training and test data come from distinct, but related, domains. 
For example, in our dataset the domain generalization challenge comes from training and test data sourced from distinct sonar camera deployments on different rivers.
%This has been shown to confound algorithms across a range of machine learning applications \cite{torralba2011unbiased,gulrajani2020search,ye2022ood,koh2021wilds}.
Crucially, as opposed to \textit{domain adaptation}, in which data from the test domain is available during model training, in domain generalization data from the test domain is considered inaccessible~\cite{muandet2013domain,blanchard2011generalizing,torralba2011unbiased}. Within computer vision this has been most extensively studied for the task of object recognition~\cite{csurka2017domain, Zhou2021DomainGI,gulrajani2020search,ye2022ood}, with a number of supporting datasets~\cite{fang2013unbiased,koh2021wilds,beery2018recognition,saenko2010adapting}. While there has been some study of domain generalization in other computer vision tasks such as semantic segmentation~\cite{hoffman2016FCNs,chen2017ROAD,zhang2017urban} and action recognition~\cite{soomro2012ucf101,weinland2006free,kuehne2011hmdb}, it has been relatively understudied in the context of MOT and counting. Some MOT and counting datasets do represent domain generalization challenges in their test sets---for example, in MOT20~\cite{dendorfer2020mot20} one of the three test locations is from a new location, and in UA-DETRAC~\cite{wen2020ua} (MOT) and WorldExpo~\cite{zhang2016data} (crowd counting) all test data is from different locations than training data. Our dataset makes it possible to evaluate generalization for both MOT and video-based counting concurrently, while providing more out-of-distribution test videos than existing options.
%(see Tab. \ref{table:datasets}).
%\sz{That said, there are difference nuances and perspectives of generalization. For example, \cite{beery2018recognition} generalized methods from domain adaptation in classification from \cite{csurka2017domain} to generalize animal detection and classification to new locations. \cite{efros2011datasetBias} showed generalization difficulties by pointing out biases in datasets, as they categorized not just what was in an image, but which dataset the image came from. \cite{hoffman2016FCNs,chen2017ROAD,zhang2017urban} looked into methods of domain adaptation for semantic segmentation tasks, though their methods were largely based on detection rather than tracking. More limited are the number of methods and datasets that generalize to different domains in multi-object tracking  \cite{csurka2017domain, Zhou2021DomainGI}. \cite{van2017devil} showed that the then-SOTA classification algorithms performed well on uniformly distributed data, but did not generalize well to long-tailed distributions. However, the natural world is long-tailed. Our differences in detection and tracking performance across each of these rivers reflects this long-tail, and our methods aim to generalize across each of the rivers. This will allow the domain to adapt to unseen rivers and scale to new rivers that the model has not been trained on.} 

%\subsubsection{Imaging Sonar} 
\noindent
{\bf Imaging Sonar.} Only a very limited amount of annotated sonar imaging data has been released. \cite{mccann2018underwater} collected 524 sonar video clips with video-level species labels, \cite{schneider2020counting} collected a small dataset of 143 sonar images to test image-level classification of fish and dolphin species, and \cite{liu_counting_2018} collected dot annotations for counting fish in 537 sonar images. Our work is the first to release detection and tracking annotations for fish in sonar, and it is several orders of magnitude larger than existing sonar datasets.

\section{Dataset}

\begin{figure}[t]
    \centering
    \includegraphics[width=\textwidth]{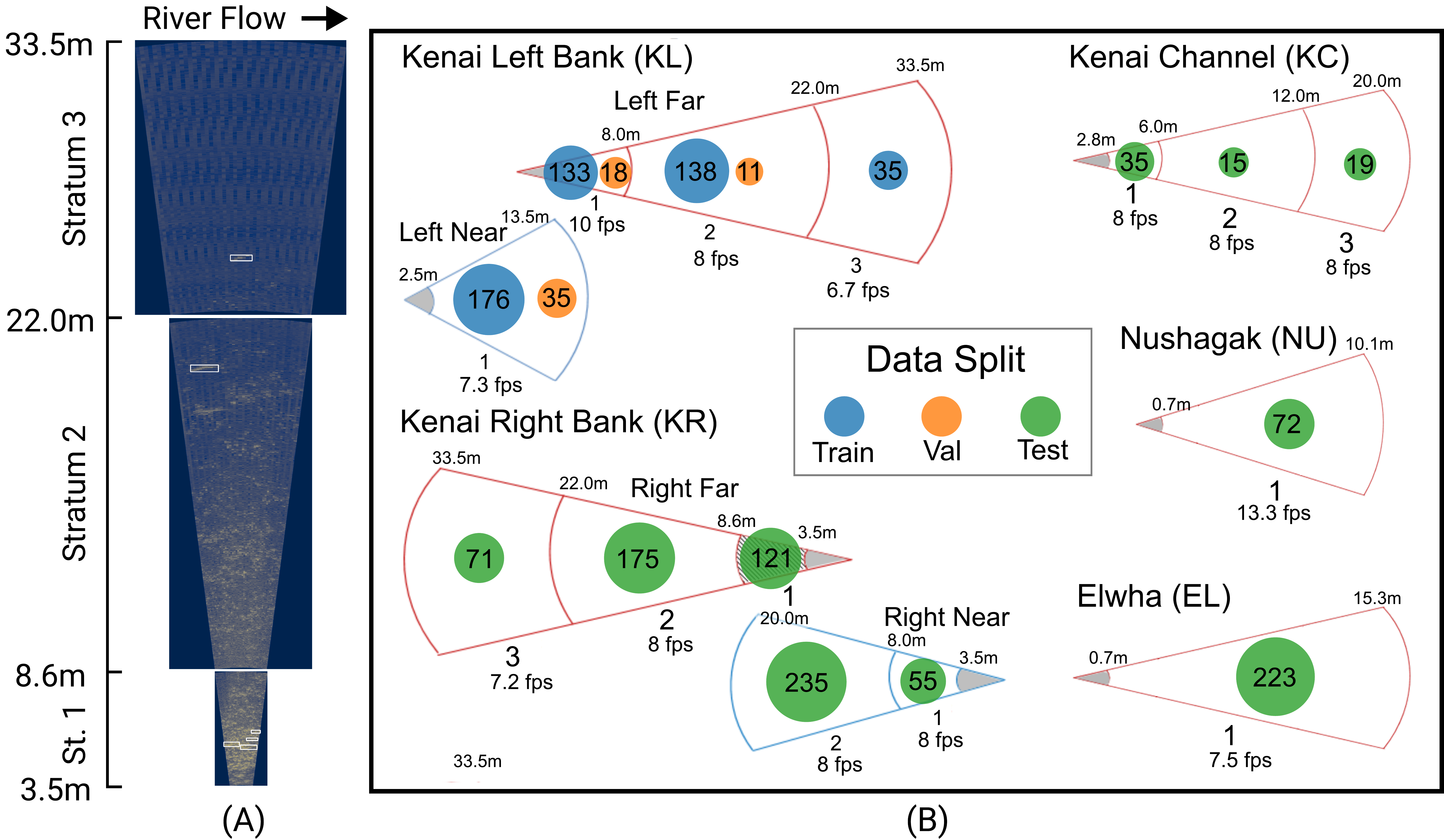}
    \caption{\textbf{Illustration of camera strata and data split.} \textbf{(A)}~Three example frames, one from each range window (``stratum'') of the far-range camera at the KR location. Sonar cameras typically cycle between multiple strata periodically. White bounding boxes are ground-truth fish locations. \textbf{(B)}~All cameras in the dataset. There are seven cameras total distributed among the five locations in the dataset, each with between one and three strata. Data split is indicated by the colored circles, and the number of training, validation, and testing sequences are indicated for each camera/stratum.}
    \label{fig:kenai}
\end{figure}

\begin{figure}[t]
    \centering
    \includegraphics[width=\textwidth]{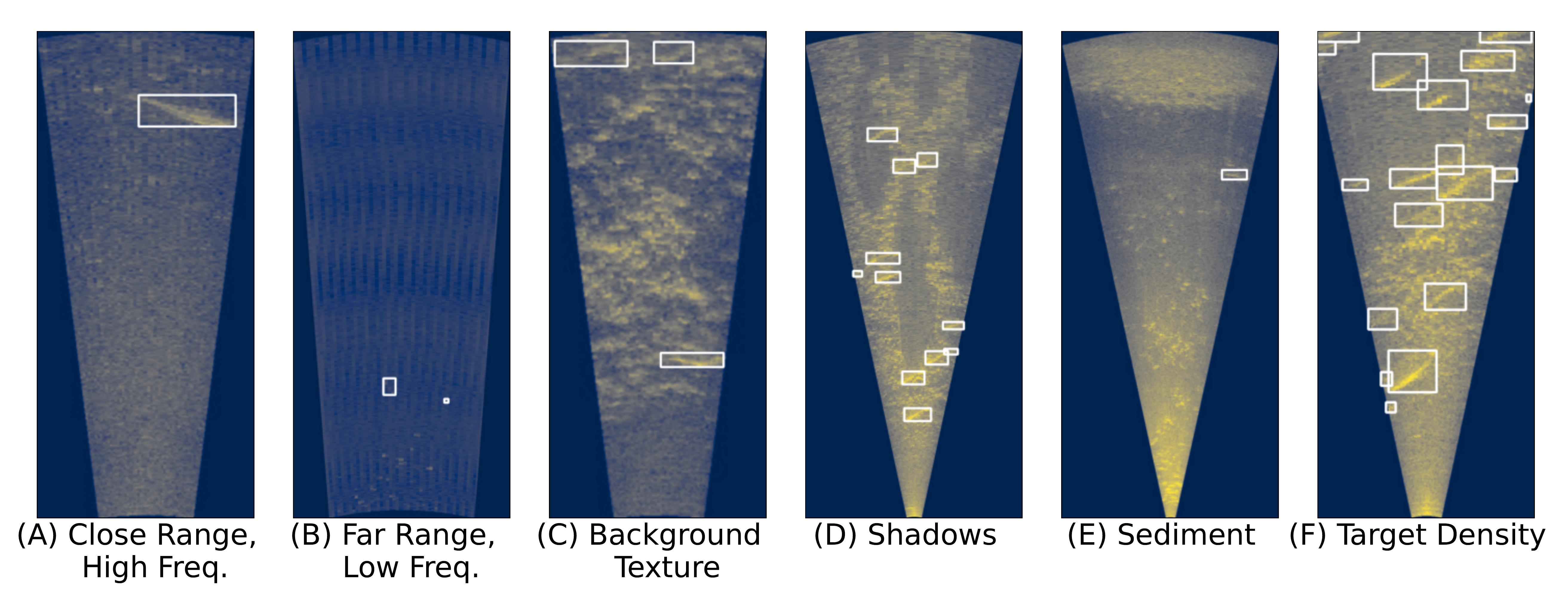}
    \caption{\textbf{Example frames and common challenges in sonar video.} Ground-truth fish locations are boxed in white. ~\textbf{(A)}~Close range, high operating frequency: ideal conditions, fish is large, visible, and well-defined. Some speckle noise is still present. \textbf{(B)}~Far range, low operating frequency: The fish are small and very coarsely defined due to scattering of sound waves at long range. \textbf{(C)}~Background texture: The riverbed is very visible, occluding fish. \textbf{(D)}~Shadows: Fish cast acoustic shadows which may occlude one another. \textbf{(E)}~Sediment: Dirt, debris, and glacial silt occlude fish. \textbf{(F)}~Target density: Dense crowds of fish, intersecting trajectories and occlusion. 
    }
    \label{fig:example}
\end{figure}

% Here we describe how we collected~(Sec. \ref{sec-collection}), annotated~(Sec. \ref{sec-annotation}), and split~(Sec. \ref{sec-split}) \datasetname, and give an overview of common challenges~(Sec. \ref{sec-challenges}).
Here we describe how we collected, annotated, and split \datasetname, and give an overview of common challenges.

%\subsection{Data Collection}
\label{sec-collection}
\noindent
{\bf Data Collection.} The dataset was curated from 2,056 hours of sonar video obtained from the Alaska Department of Fish and Game, the U.S. National Marine Fisheries Service, the U.S. National Park Service, and the Lower Elwha Klallam Tribe. It contains video from five distinct locations which we use to study out-of-sample performance: three locations on the Kenai River in Alaska, which we refer to as \textbf{KL} (\textbf{K}enai \textbf{L}eft Bank), \textbf{KR} (\textbf{K}enai \textbf{R}ight Bank), and \textbf{KC} (\textbf{K}enai \textbf{C}hannel); one location, \textbf{NU}, on the \textbf{Nu}shagak River in Alaska; and one location, \textbf{EL}, on the \textbf{El}wha River in Washington (sonar configurations shown in  Fig.~\ref{fig:kenai}). 

The data had already been analyzed by experts (``manually marked'') to obtain fish counts. Most of the video contained no fish. Since our focus is on detection, tracking, and counting, we used the manual markings to extract shorter 200--300 frame video clips known to contain fish. If any of these clips overlapped, we merged them into one longer clip. In total we extracted 1,233 clips from the Kenai River, containing 4,300 fish; 262 clips from the Elwha River, containing 884 fish; and 72 clips from the Nushagak River, containing 3,070 fish.

%\subsection{Annotation}
\label{sec-annotation}
\noindent
{\bf Annotation.} We hired a third-party annotation service to collect multiple-object tracking annotations for all fish in the extracted clips. Annotators were provided with the raw video clips and instructed to box all visible fish tightly using the vatic.js GUI~\cite{bolkensteyn_dbolkensteynvaticjs_2020}. For any stationary fish, they were required to annotate every fifth frame, and we interpolated between those annotations in the intermediate frames; for all other tracks, all bounding boxes in all frames were annotated manually. The annotation service had their own internal quality-management procedures whereby multiple annotators inspected each clip before it was finalized. In total, 515,933 bounding boxes for 8,254 fish tracks were collected in 527,215 frames from seven different cameras.

%While not our main focus, some manual markings contained ground truth fish length measurements, which we associate with annotated tracks and include in the dataset release. We discuss a simple baseline method for size prediction in the supplemental material. 
% As additional quality control, we ran an experiment comparing our own ``expert'' annotations (i.e. tracking annotations created by all team members using the same tool) on one randomly selected clip from each camera in the dataset. We found that \jk{results from this experiment}. This experiment additionally motivated our choice of IoU threshold in computing detection and tracking metrics (see Section \ref{sec-metrics}).
% \jk{If we want to say more about the ground-truth count and/or length annotations, we can do so here. TODO: It is probably worth a second pass at those counts to make sure they are actually as bad as we think (59\% discrepancy between annotators and biologists)...}

%\subsection{Data Split}
\label{sec-split}
\noindent
{\bf Data Split.} We designed a dataset splitting protocol that allows us to study generalization to new locations, a known challenge for current computer vision methods \cite{wilds,beery2018recognition}. Our test data comes from deployment locations never seen during training or validation. We chose KL as our training and validation location due to its sufficient size for model training. Data from this location spans 16 days in total. We selected one of these days at random to hold out as a validation set, and the other 15 days serve as our training set. This gives us 162,680 training images containing 1,762 tracks and 132,220 bounding boxes, and 30,518 validation images containing 207 tracks and 18,565 bounding boxes.

The other locations (KR, KC, NU, and EL) serve as test sets for evaluating generalization performance under different conditions. These locations cover a range of generalization scenarios that may be faced in the real world: KR includes a new camera deployment on the same body of water; KC includes a new deployment in a nearby, but separate, body of water; NU and EL include deployments on new rivers in different geographic regions with different species distributions. The data split is illustrated in Fig. \ref{fig:kenai}. In total the dataset contains 334,017 test images with 6,285 tracks and 365,148 bounding boxes. For all experiments we report results on all test locations individually.

%\subsection{Challenges}
\label{sec-challenges}
\noindent
{\bf Challenges.} We have identified a number of challenges inherent to detecting and tracking fish using sonar, which we have illustrated in Fig. \ref{fig:example}. Some of these challenges are constant across all data in this domain (e.g. speckle noise and shadows), while some vary across locations due to hardware settings or environmental factors, presenting generalization challenges (e.g. presence of sediment, riverbed shape and texture, and hardware operating frequency). More details on the causes of these challenges can be found in the supplemental material.

\section{Metrics}
\label{sec-metrics}

\subsection{Counting Protocol}
\label{sec-protocol}

We follow the counting procedure used by field technicians when counting fish in sonar video~\cite{kenaiplan}. A vertical line-of-interest (LOI) is drawn in the middle of the frame, and a fish is considered to have moved left or right if its trajectory start and end positions are on different sides of the LOI. Note that not every fish in a clip will cross the LOI. Some fish are stationary throughout the entire clip, while others enter and exit on the same side of the frame without crossing the LOI. These fish are excluded from the count totals, which matches the protocol used by the field technicians.

\subsection{Counting Metric}

In the target application, fish-counting error is measured as the sum of upstream and downstream counting errors, and error is normalized separately at each river to account for variations in fish abundance. We classify direction of movement as ``left'' or ``right'' rather than ``upstream'' or ``downstream'' to make our system agnostic to the orientation of the camera. Based on this we define the absolute counting error for the $i$th video clip, $E_i$, as the sum of the absolute left and right (i.e. upstream and downstream) counting errors:
\begin{eqnarray}
    E_i = | z_{left_i} - \hat{z}_{left_i} | + | z_{right_i} - \hat{z}_{right_i} |
\end{eqnarray}
where $z_i$ and $\hat{z}_i$ are the predicted and ground-truth counts for clip $i$, respectively. Overall counting error is reported as normalized Mean Absolute Error (nMAE):
\begin{eqnarray}
    nMAE = \frac{\frac{1}{N}\sum_{i=1}^{N} E_i}{\frac{1}{N}\sum_{i=1}^{N} \hat{z}_i} = \frac{\sum_{i=1}^{N} E_i}{\sum_{i=1}^{N} \hat{z}_i}
\end{eqnarray}
where $N$ is the number of video clips at a given location and $\hat{z}_i = \hat{z}_{left_i} + \hat{z}_{right_i}$ is the ground-truth count for clip $i$.

%We choose a normalized metric to account for large variations in the number of tracks per video clip across different locations.
% This is compared to \textit{per-clip} normalization, which would give equal weight to each \textit{clip} at a given location, 
% clips are arbitrarily generated and have varying number of fish. A metric that normalizes per-clip would be skewed by these factors. 
% Compared to \textit{per-clip} normalization, this has the advantage of reducing the influence of small errors in sparse clips, but the disadvantage of giving more weight to denser clips within a given location. 
%Given that daily error statistics are sufficient in the target application, this tradeoff is acceptable.
%We do this because in the target application, counts are computed per-day, i.e. over a set of several hours of video, whereas our clips are arbitrarily chosen and are much shorter (30 seconds -- 1 minute long)

%We do this because a ``clip'' in our dataset is not relevant to the target application, as our clips are much shorter in duration than the original videos; thus, normalizing pier-clip would not make sense for the target application and would be overly influenced by fish in sparse clips.

We choose a metric that is normalized \textit{per location}, giving \textit{equal weight to each fish} at a given location regardless of variance in fish density across clips at that location.
We do this because video clips in \datasetname are arbitrarily generated and are much shorter in duration than videos in the target application, thus \textit{per-clip} normalization---i.e. giving equal weight to each \textit{clip}---would not be appropriate.
%and have varying number of fish. A metric that normalizes \textit{per-clip} would be overly influenced by fish in sparse clips.
% would given extra weight to fish in sparse clips
% bias counting performance toward clips with few fish.
% skewed by these factors in a way that biases
%makes counting performance not easily interpretable.
%(Imagine an algorithm that only does well on clips with 0 or 1 fish vs an algorithm that does well on clips with many fish, we prefer the algorithm that matches the ground truth count fish most closely, as opposed to the algorithm that does best on the average clip.)
% We choose a metric that is normalized \textit{per location} to account for large variations in the number of tracks per video clip across different locations.
We consider this the main metric for the dataset, since it is the most important in the target application. Achieving at most 10\% counting error on a river would make an algorithm on par with human experts and feasible for augmenting counting in the field~\cite{kenaiplan}.

\subsection{Detection and Tracking Metrics}

In addition to counting, \datasetname has been annotated to measure detection and tracking performance as well. For our detection metric we choose the PascalVOC~\cite{everingham2010pascal} evaluation of mean Average Precision with IoU $\geq 0.5$ (AP50). For tracking, we report the CLEAR~\cite{bernardin2008evaluating}, IDF1~\cite{ristani2016performance}, and HOTA~\cite{luiten2021hota} metrics.
%the two most common MOT metrics, MOTA~\cite{bernardin2008evaluating} and IDF1~\cite{ristani2016performance}.

The CLEAR MOT metrics~\cite{bernardin2008evaluating} are computed \textit{per-frame} by matching detections from predicted tracks with ground-truth detections. This matching allows the number of true positive ($TP$), false positive ($FP$), and false negative ($FN$) detections to be computed using an IoU threshold between predicted and ground-truth boxes. Recall (\textit{CLR\_Re}) and precision (\textit{CLR\_Pr}) are defined as normal according to the number of TPs, FPs, and FNs, and Multiple Object Tracking Accuracy (\textit{MOTA}) is defined as:
\begin{equation}
    MOTA = \frac{TP - FP - IDSW}{TP + FN}
\end{equation}

\noindent Where $IDSW$ is the sum of all ``ID switches'', i.e. the number of times a predicted track changes its matched ground-truth track and vice versa. In practice, $IDSW \ll TP$ and MOTA becomes predominantly a measure of detection quality in the tracks. Since this scoring occurs per-frame, MOTA does not capture long-term tracking performance.

In contrast, IDF1~\cite{ristani2016performance} first computes a \textit{global} (per-video-clip) \textit{track} matching, i.e. a bipartite matching between all ground-truth and predicted tracks. From this matching, ID true positives ($IDTP$), false positives ($IDFP$), and false negatives ($IDFN$) are computed, and the IDF1 score is defined as the harmonic mean of ID Precision (IDP) and ID Recall (IDR):
\begin{equation}
    IDF1 =2 \cdot \frac{IDP \times IDR}{IDP + IDR},~IDP = \frac{IDTP}{IDTP + IDFP},~IDR = \frac{IDTP}{IDTP + IDFN}
\end{equation}

Since this matching is restricted to be static for the length of the clip, IDF1 is a measure of long-term tracking performance.

HOTA~\cite{luiten2021hota} is designed to measure both short-term and long-term tracking performance. It is the geometric mean of a detection score (\textit{DetA}) and an association score (\textit{AssA}), each defined as the Jaccard index of detection/association TPs, FPs, and FNs. Thus it can be easily decomposed into detection/association precision and recall (\textit{DetRe} and \textit{DetPr}, and \textit{AssRe} and \textit{AssPr}, respectively). In this formulation, AssRe is inversely correlated with the number of track splits, while AssPr is inversely correlated with the number of track merges.

\section{Experiments}
\label{sec-experiments}

We evaluate state-of-the-art methods on \datasetname to provide a baseline for future work and give insight into the generalization challenges of object detection, multiple-object tracking, and counting. In Sec.~\ref{sec-baseline} we propose a tracking-by-detection approach to fish counting which allows us to evaluate each of these tasks, and study its performance. In Sec.~\ref{sec-upperbounds} we perform ablation studies and  investigate the upper bounds of this approach and its generalization capabilities. In Sec.~\ref{sec-baselinepp} we introduce an improved baseline method to address these challenges, establish the state of the art on \datasetname, and discuss remaining challenges. 
% We note that this is not the only way to perform counting and future work may include methods for directly predicting counts without MOT.
% We choose this approach because it is interpretable, matches the procedure used in the target application, and allows us to extract additional meaningful information (i.e. size and direction of motion) from the predicted tracks. We note that this is not the only way to perform counting and future work may include methods for directly predicting counts without MOT.

\subsection{Baseline}
\label{sec-baseline}

\begin{figure}[t]
    \centering
    \includegraphics[width=\textwidth]{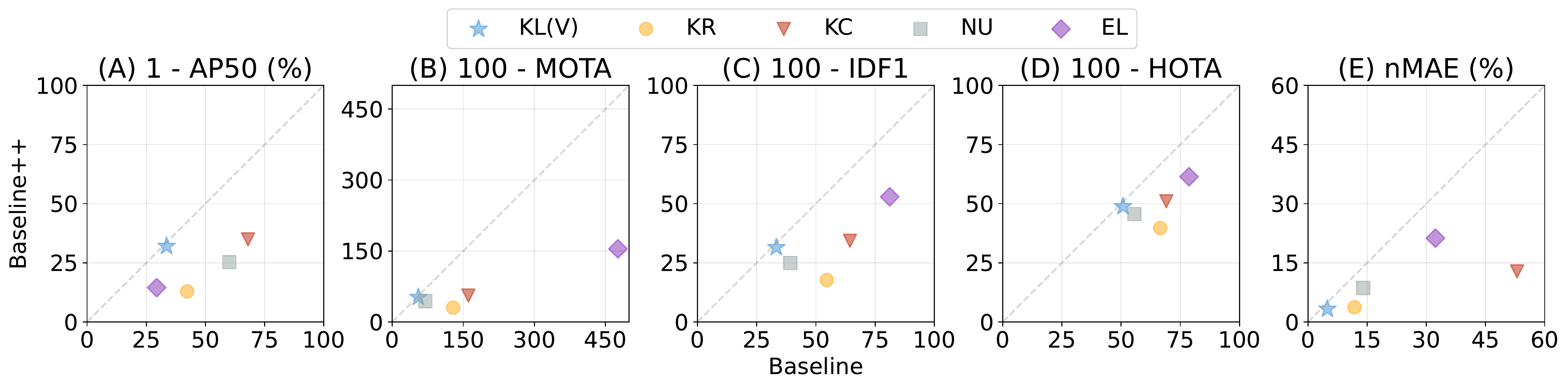}
    \caption{\textbf{Baseline (Sec. \ref{sec-baseline}) and Baseline++ (Sec. \ref{sec-baselinepp}) results on \datasetname.} All results are displayed in terms of error, i.e. the lower-left of each plot represents the best performance. Baseline++ improves performance on all tasks across all locations and reduces generalization gaps between the validation location and testing locations. Generalization challenges remain, most notably at KC and EL.}
    \label{fig:tab2_replacement}
\end{figure}

Our baseline method uses the YOLOv5~\cite{glenn_jocher_2022_6222936} object detector and SORT~\cite{bewley2016simple} tracker. Trajectories are then analyzed as described in Sec. \ref{sec-protocol} to predict counts.

\begin{figure}[t]
    \centering
    \includegraphics[width=\textwidth]{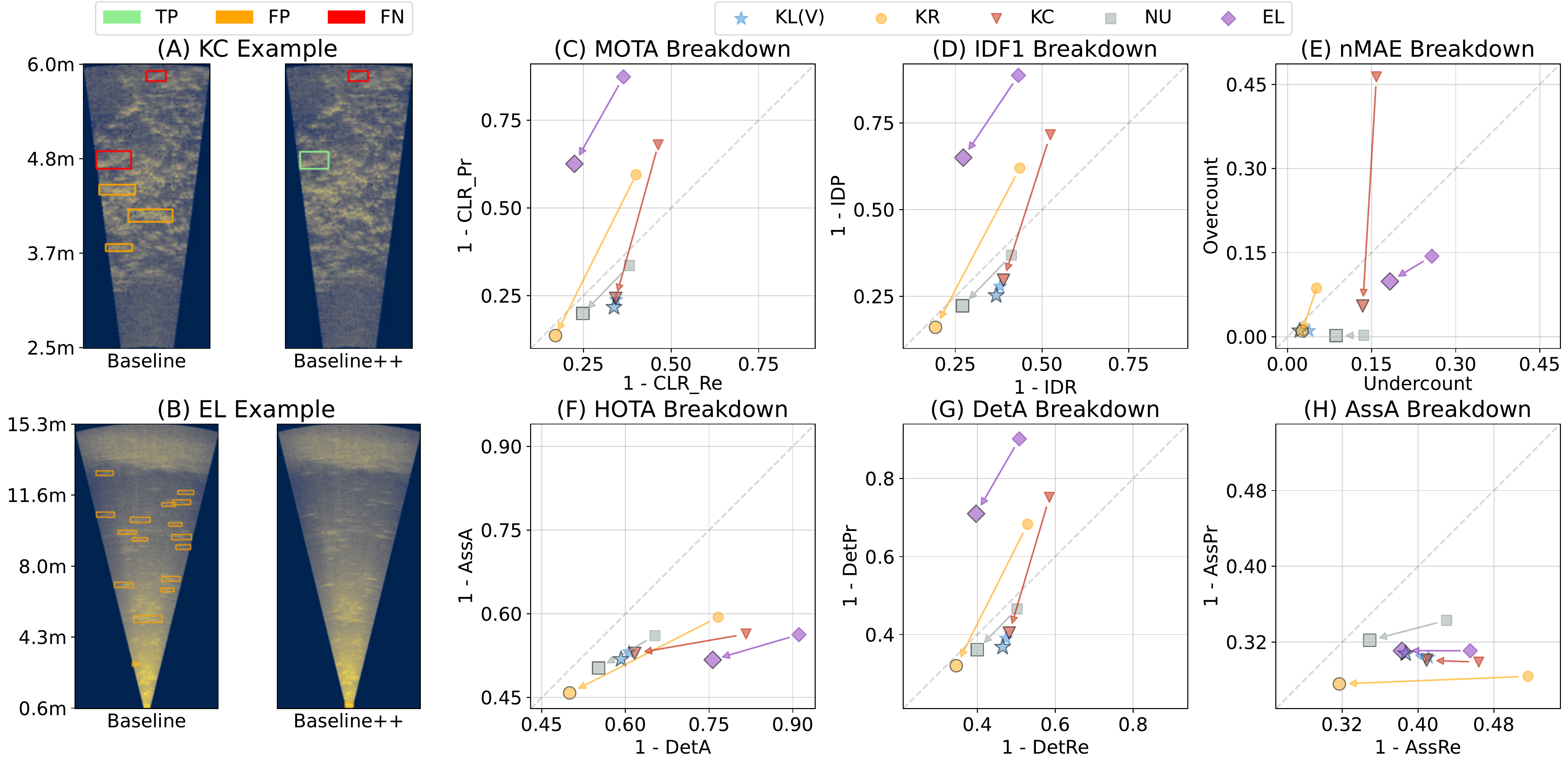}%{figs/baseline_breakdown_v4.pdf}
    \caption{\textbf{Baseline and Baseline++ error analysis and comparison.}
    \textbf{(A)--(B)}~Example frames at the KC and EL locations. Note the large reduction in FP and FN detections between the two methods.
    \textbf{(C)--(H)}~Breakdowns of MOTA, IDF1, HOTA, and nMAE into component submetrics. Arrows point from Baseline results to Baseline++ results; Baseline++ markers are larger and have a black edge. All results are displayed in terms of error. 
    This breakdown shows that the low tracking scores of our baseline method at out-of-distribution locations are predominantly due to FP detections (i.e. low detection precision) that cause low CLR\_Pr (C), low IDP (D), and low DetPr leading to low DetA (F)--(G). The proposed Baseline++ method succesfully targets a large portion of these FPs, improving all tracking metrics and submetrics.
    \textbf{(E)}~nMAE decomposed into undercounting and overcounting errors, normalized by ground-truth counts. Baseline++ significantly reduces both types of errors.
    }
    \label{fig:baseline_breakdown}
\end{figure}

We chose YOLOv5 after an initial architecture search. These experiments as well as training settings are included in the supplemental material. We chose SORT because it has shown to be a popular and robust tracker across a range of applications, recently achieving state-of-the-art performance on several MOT datasets with minor modifications~\cite{zhang2021bytetrack}. It performs tracking using a motion model based on the Kalman filter \cite{kalman_new_1960} without using appearance information for association. We verified our hypothesis that appearance features are not a strong signal for association in \datasetname by training a visual re-identification model that performed poorly on our validation set. More details of this experiment are provided in the supplemental material. See Figs.~\ref{fig:tab2_replacement} and \ref{fig:baseline_breakdown} for our performance on the various locations with this baseline method. A tabular version of these results is included in the supplemental material.

% \subsubsection{Analysis}

% This breakdown shows that the the low MOTA and DetA scores of our baseline method are predominantly due to FP detections, and these FPs propogate to cause low IDP and HOTA as well. The proposed Baseline++ method succesfully targets a large portion of these FPs, improving all tracking metrics and submetrics.
    % (E):~nMAE decomposed into undercounting and overcounting errors, normalized by ground truth counts. Baseline++ significantly reduces both types of errors.
    % Top row is Baseline method (Sec. \ref{sec-baseline}), bottom row is Baseline++ (Sec. \ref{sec-baselinepp}). (A)--(B):~example frames at the KC and EL locations, with the background-subtracted version of the same frames shown in the bottom row. Note the large reduction in FP and FN detections between the two methods. (C)--(D):~breakdowns of MOTA and IDF1 results into their component submetrics. Gray curves in (D) are IDF1 isocurves. This breakdown shows that the the low MOTA scores of our baseline method are predominantly due to FP detections, and these FPs propogate to cause low IDP as well. The proposed Baseline++ method succesfully targets a large portion of these FPs, improving all tracking metrics and submetrics. (E):~nMAE decomposed into undercounting and overcounting errors, normalized by ground truth counts. Baseline++ significantly reduces both types of errors. 

\noindent
\textbf{Analysis.} Our baseline method performs on par with human experts at the location where it is trained, with a validation counting error of less than 5\%. Counting generalization performance is best at KR, which matches intuition given that this data is sourced from a nearby location to the training set (KL). However, generalization performance at the other locations is quite poor. We examine the two most challenging locations, KC (53\% nMAE) and EL (32.3\% nMAE), in Fig.~\ref{fig:baseline_breakdown}. At KC, false positive (FP) and false negative (FN) detections are caused by the presence of complex background information (Fig.~\ref{fig:baseline_breakdown}A), while at EL the errors are overwhelmingly FPs caused by sediment and other noise resulting from the camera's very large range window (Fig.~\ref{fig:baseline_breakdown}B).

We can see the direct impact of these detection errors on our tracking and counting metrics in Fig.~\ref{fig:baseline_breakdown}C--H. At both locations, we see that abundant FPs cause very low CLR Precision (Fig.~\ref{fig:baseline_breakdown}C), ID Precision (Fig.~\ref{fig:baseline_breakdown}D), and Detection Precision (Fig.~\ref{fig:baseline_breakdown}G), negatively impacting MOTA, IDF1, and HOTA, respectively. Interestingly, while FPs cause overcounting errors at KC, the majority of counting errors at EL are actually due to undercounting (Fig.~\ref{fig:baseline_breakdown}E). From manual inspection we diagnosed that this is often caused by TP tracks merging with hallucinated detections, causing the loss of a track before it has the chance to cross the counting line. These challenges demonstrate why counting, in addition to detection and tracking, is an important metric for researchers to consider.

% At KC, there is a high ratio of both FPs and FNs to TPs (Fig. \ref{fig:baseline_breakdown}C), causing a very low MOTA (-60.8). This affects both IDR and IDP, with KC having the lowest IDR of any location. This is reflected in the HOTA breakdown (Fig. \ref{fig:baseline_breakdown}F---H) as well: the low HOTA score at KC is caused by a poor detection score (Fig. \ref{fig:baseline_breakdown}F), which is in turn caused by low detection precision (Fig. \ref{fig:baseline_breakdown}G).
% In the case of EL, false positives outnumber true positives by nearly 7x (Fig. \ref{fig:baseline_breakdown}C), causing a very low MOTA score (-376.7). This also results in the lowest IDP score of any location (11.3\%), due to the hallucination of additional tracks as well as the merging of TP tracks with hallucinated detections. We see this trend reflected in the HOTA metrics as well, with EL achieving the worst DetA and DetPr performance of any location.

Our baseline results indicate that (1) there are indeed generalization challenges in this domain, (2) they appear to be largely caused by location-specific environmental changes, and (3) these challenges affect all three tasks of interest: detection, tracking, and counting. Further, the low counting error on the validation set indicates that our overall tracking-by-detection approach is feasible at in-distribution locations and that the key challenges are in generalization. To verify this, in the next section we perform ablation studies and examine the upper-bound generalization capabilities of the proposed tracking-by-detection baseline, and use these results to motivate an improved baseline in Sec. \ref{sec-baselinepp}.

\subsection{Ablation Study and Generalization Upper Bounds}
\label{sec-upperbounds}

\begin{figure}[t]
    \centering
    \includegraphics[width=\textwidth]{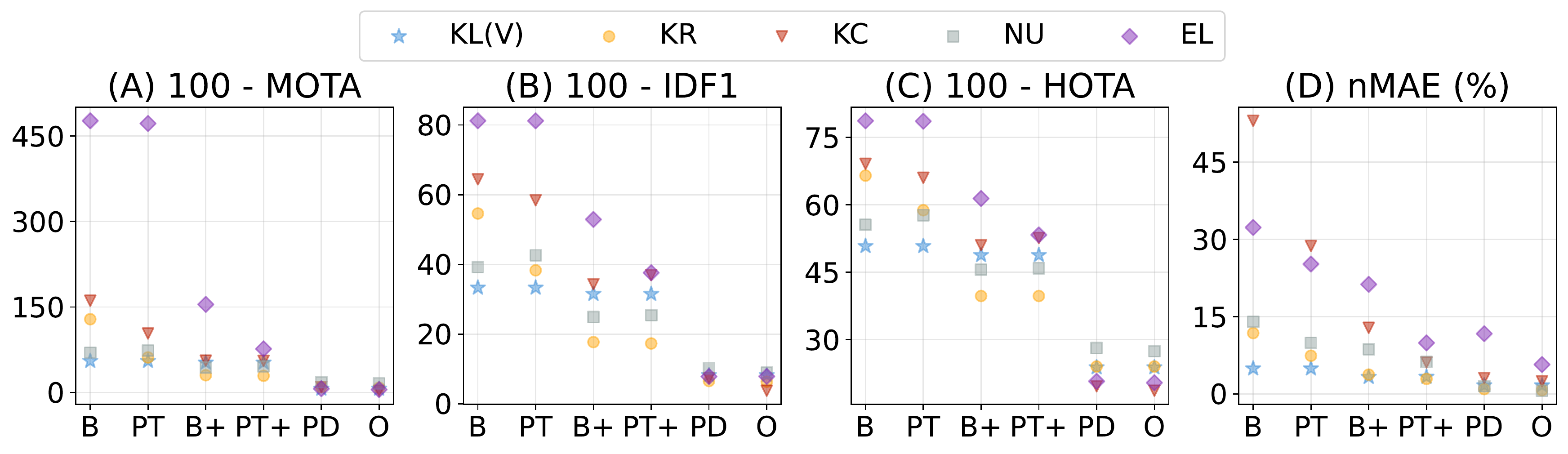}
    \caption{\textbf{Baseline and upper bound tracking and counting performance.}
    Baseline methods are compared to methods that use various amounts of ground-truth data. 
    X axis key: \textbf{B}aseline, \textbf{P}erfect \textbf{T}racker, \textbf{B}aseline\textbf{+}+, \textbf{P}erfect \textbf{T}racker\textbf{+}+, \textbf{P}erfect \textbf{D}etector, \textbf{O}racle.
    Note the performance difference between Baseline and Perfect Detector: there is a large performance boost when using ground-truth detections. Using a tracker that can perfectly generalize to different locations improves performance as well, but not as much as the Perfect Detector. 
    %B+ improves detection and thus overall performance with respect to B. %this is emphasized in Fig 3 and 4 - I think this Fig should emphasize the upper bounds analysis
    }
    \label{fig:generalization}
\end{figure}

To evaluate generalization potential, we perform a set of ``reversed'' ablation studies and compare our baseline results with three different upper bounds that utilize different types of ground-truth information (Fig.~\ref{fig:generalization}):
% \sara{I said this earlier too, but I love this as a tool for more detailed understanding of performance breakdowns}

%\begin{enumerate}
    %\item 
    \textbf{1. Perfect Tracker Generalization.} We use the detections from our baseline detector, but we fit the tracker hyperparameters directly to each test set based on counting performance. This results in five different trackers, one for each test set, with the best possible tracker parameter settings at each location. This gives an upper bound for the counting performance of a multiple-object tracker which can perfectly generalize to new locations. 
    
    %\item 
    \textbf{2. Perfect Detector.} This model takes \textit{ground-truth} detections as input. The tracker hyperparameters are fit to counting performance on the validation set as in our baseline. This gives an upper bound for the counting performance of our tracker when given perfect detections.
    
    %\item 
    \textbf{3. Oracle.} This model combines the first two upper bounds. It takes ground-truth detections as input and fits the tracker hyperparameters to each test set. This gives us the overall upper bound on counting performance for our baseline approach given perfect detections and perfect tracker generalization.
%\end{enumerate}

% \subsubsection{Analysis} 
\noindent
\textbf{Analysis.} The most apparent result is the very strong performance at all locations given a Perfect Detector. In most cases, with perfect detections our tracker achieves near-Oracle performance and generalizes well without modification. Only one location (EL) shows significant further improvements in counting error in the Oracle method compared to the Perfect Detector method. Meanwhile, Perfect Tracker Generalization does improve performance in most cases, but not as much as the Perfect Detector. This indicates that the proposed motion-based tracking approach is indeed feasible but is dependent upon a strong detector with strong generalization capabilities. Therefore the most effective improvement to overall system-generalization performance appears to be improving the generalization capabilities of the detector, which we address in the next section.

\subsection{Baseline++}
\label{sec-baselinepp}

% Given our previous ablation studies on which changes can lead to the largest improvements, i
Given the results from our upper bound analysis, we implemented an improved baseline method, ``Baseline++,'' with the primary goal of improving object-detection generalization performance.
% In this section we introduce an improved baseline method, ``Baseline++,'' establishing state-of-the-art performance on \datasetname. Given the results from our upper bound analysis, our primary goal was improving object-detection generalization performance. 
We noticed that the background (1) varies significantly across locations and (2) occludes fish (see Fig.~\ref{fig:example} for frame examples). Thus, we appended two additional channels to our image input: (1) a background-subtracted version of each frame, where the background for each clip is obtained by averaging all frames, and (2) the difference between each background-subtracted frame and its preceding frame, to capture motion information. Example frames illustrating these transformations are included in the supplemental material. We trained a new detector and tracker with this input in the same way as the baseline model. Results are shown in Figs.~\ref{fig:tab2_replacement}, \ref{fig:baseline_breakdown}, and \ref{fig:generalization}.

%\subsubsection{Analysis} 
 %In Figs.~\ref{fig:tab2_replacement} and \ref{fig:baseline_breakdown} we compare the Baseline and Baseline++ methods. 
\noindent
{\bf Analysis.} In Fig.~\ref{fig:tab2_replacement}, we see that our Baseline++ method leads to modest improvements on our validation set (+1.6 AP50, -1.7\% nMAE), but \textit{significant} improvements in generalization performance (e.g. -40.2\% nMAE at KC). 
In Fig.~\ref{fig:baseline_breakdown}, we dissect these improvements by looking at: (A)--(B) two example frames from KC and EL, and (C)--(H) breakdowns of tracking and counting metrics across all locations. We see the efficacy of simple background subtraction as a generalization mechanism, helping significantly reduce the number of FPs and FNs in the example KC and EL frames as well as at all test locations. We also see evidence of some outstanding issues: one FN remains in the KC example due to a small, stationary fish that appears to have been removed by the background-subtraction routine, and one FP remains in the EL example due to noise near the transducer. IDR now lags behind IDP at all locations except for EL, indicating that most remaining tracking problems are track splits causing undercounting errors. These trends are also indicated in the HOTA decompositions (Fig.~\ref{fig:baseline_breakdown}F--H), which show that DetRe now lags behind DetPr at all locations except EL, and while AssRe has improved (i.e. track splits have been reduced), it is still lower than AssPr.  
%\jk{Final sentence about where future improvements should focus?}

% In Fig. \ref{fig:baseline_breakdown} (bottom) we show: (A)--(B) the same two example frames from the baseline evaluation (displayed is the background-subtracted channel), with output from the Baseline++ model overlaid, and (C)--(D) the same tracking and counting breakdowns across all locations as in Sec. \ref{sec-baseline}. Baseline++ performance is also shown in Table \ref{table:overall_results} and Fig. \ref{fig:generalization} for reference with detection and upper-bound performance. With these enhancements we see modest improvements on our validation set (+1.6 AP50, -1.7\% nMAE), but significant improvements in generalization performance (e.g. -40.2\% nMAE at KC). In Fig. \ref{fig:baseline_breakdown} we see the efficacy of simple background subtraction as a generalization mechanism, helping significantly reduce the number of FPs and FNs in the example KC and EL frames as well as at all test locations. We also see evidence of some outstanding issues: one false negative remains at the KC frame due to a small, stationary fish that appears to have been removed by the background-subtraction routine; and one false positive remains at the EL frame due to oscillations near the transducer. IDR now lags behind IDP at all locations except for the Elwha, indicating that remaining tracking problems are primarily track splits, causing undercounting errors.

\section{Conclusions}
\label{discussion}

We present the \datasetfullname, a natural-world sonar video dataset that allows us to study object detection, multiple-object tracking, and counting under challenging real-world domain shifts. Due to the visual qualities of the source domain of river-based sonar, the dataset poses challenges to existing methods developed primarily for urban environments and provides a benchmark for video-based counting in the wild, a task that lacks supporting benchmarks.

%% Moved from previous section - do we want to keep any?
% \tiffany{the content follows could be delayed to discussion, here it just sounds like we do not have time to do X,Y,Z. We could also not mention things we did no do...} While these performance improvements are encouraging, the remaining challenges indicate that there is still significant room for improvement for generalizing state-of-the-art detection, tracking, and counting algorithms. The \jk{DatasetName} dataset can provide a challenging benchmark for the development of improved algorithms for these tasks. While we have pursued system improvements through improved detection generalization, there are also opportunities for approaches which target tracker generalization or robust tracker performance given noisy detections. One additional path forward for the community is to explore the impact of utilizing unlabeled data for unsupervised domain adaptation or self-supervised pretraining on unseen locations, which we make possible by releasing an additional unlabeled batch of sonar data from all test locations. We leave this as future work.

Our experiments show that there is still significant room for improvement in the generalization performance of tracking and counting algorithms. There are also opportunities to improve tracker generalization and making trackers more robust to noisy detections. Robust algorithms that work across the range of generalization challenges in \datasetname will certainly be impactful in other domains, and we hope that our dataset will provide a useful testing ground for the computer vision community to push forward progress on these tasks. High-performing methods would enable sonar-based fish counting to scale globally and have real-world impact in managing some of the world's most sensitive and valuable ecosystems. 

In the future, the dataset will be expanded to include additional input formats, locations, and species. One additional path forward for the community is to explore the impact of utilizing unlabeled data for unsupervised domain adaptation or self-supervised pretraining on unseen locations, which we plan to make possible by releasing additional unlabeled data from all test locations.

\noindent
\textbf{Acknowledgements.} 
% We are grateful to Amazon Web Services for a gift to Trout Unlimited that supported data annotations, computational and storage costs, and to the Resnick Sustainability Institute at Caltech for funding to SB and PP. An NSF Fellowship supported SB. EY, JK and SD volunteered their time. GVH was supported by the Macaulay Library at Cornell University. We are grateful to George Pess and Oleksandr Stefankiv (Northwest Fisheries Science Center), James Miller, Carl Pfisterer, Dawn Wilburn, Brandon Key, and Suzanne Maxwell (Alaska Department of Fish and Game), Dave Kajtaniak (California Department of Fish and Wildlife) and Dean Finnerty (Trout Unlimited's Wild Steelhead Project) for collecting and annotating the dataset, for feedback during the project, and for encouragement and moral support.
%%% ALT - including the others below
We are grateful to AWS for a gift to Trout Unlimited (TU) that supported data annotations, computational and storage costs, and to the Resnick Sustainability Institute at Caltech for funding to SB and PP. An NSF Fellowship supported SB. JK, SD, and EY volunteered their time. GVH was supported by the Macaulay Library at Cornell University. For collecting the dataset, and for feedback, encouragement, and moral support, we are grateful to: George Pess and Oleksandr Stefankiv (Northwest Fisheries Science Center); James Miller, Carl Pfisterer, Dawn Wilburn, Brandon Key, Suzanne Maxwell, Gregory Buck, April Faulkner, and Jordan Head (Alaska Department of Fish and Game); Dave Kajtaniak and Michael Sparkman (California Department of Fish and Wildlife); Dean Finnerty (TU's Wild Steelhead Project); and Keith Denton, Mike McHenry, and the Lower Elwha Klallam Tribe.

%% From workshop paper %%
% We would like to thank Amazon Web Services for funding and compute resources for this project;
% George Pess and Oleksandr Stefankiv from the National Marine Fisheries Service; James Miller,
% Suzanne Maxwell, Brandon Key, Carl Pfisterer, Gregory Buck, April Faulkner, and Jordan Head
% from the Alaska Department of Fish and Game; Michael Sparkman and David Kajtaniak from the
% California Department of Fish and Wildlife; Dean Finnerty from Trout Unlimited; as well as Keith
% Denton, Mike McHenry, and the Lower Elwha Klallam Tribe.

\clearpage
% ---- Bibliography ----
%
% BibTeX users should specify bibliography style 'splncs04'.
% References will then be sorted and formatted in the correct style.
%
\bibliographystyle{splncs04}
\bibliography{bib/behavior,bib/counting,bib/domain_adapt,bib/fgvc,bib/other,bib/reid,bib/sdm,bib/tracking}

%%%% -----Supplementary------ %%%%%%%

\title{The \datasetfullname: Supplementary Material} % Replace with your title

% INITIAL SUBMISSION 
%\begin{comment}
\titlerunning{The \datasetfullname: Supplementary Material}
% If the paper title is too long for the running head, you can set
% an abbreviated paper title here
%
\author{
Justin Kay\inst{1,5} \and
Peter Kulits\inst{1} \and
Suzanne Stathatos\inst{1} \and
Siqi Deng\inst{2} \and
Erik Young\inst{3} \and
Sara Beery\inst{1} \and
Grant Van Horn\inst{4}\index{Van Horn, Grant} \and
Pietro Perona\inst{1,2}
}
\authorrunning{J. Kay et al.}
% First names are abbreviated in the running head.
% If there are more than two authors, 'et al.' is used.
%
\institute{
California Institute of Technology \samelineand
\inst{2}\enspace AWS AI Labs \samelineand
\inst{3}\enspace Trout Unlimited \and
Cornell University \samelineand
\inst{5}\enspace Ai.Fish
}
%\end{comment}
%******************
\maketitle

\section{Imaging Sonar}

\begin{figure}[t]
    \centering
    \includegraphics[width=\textwidth]{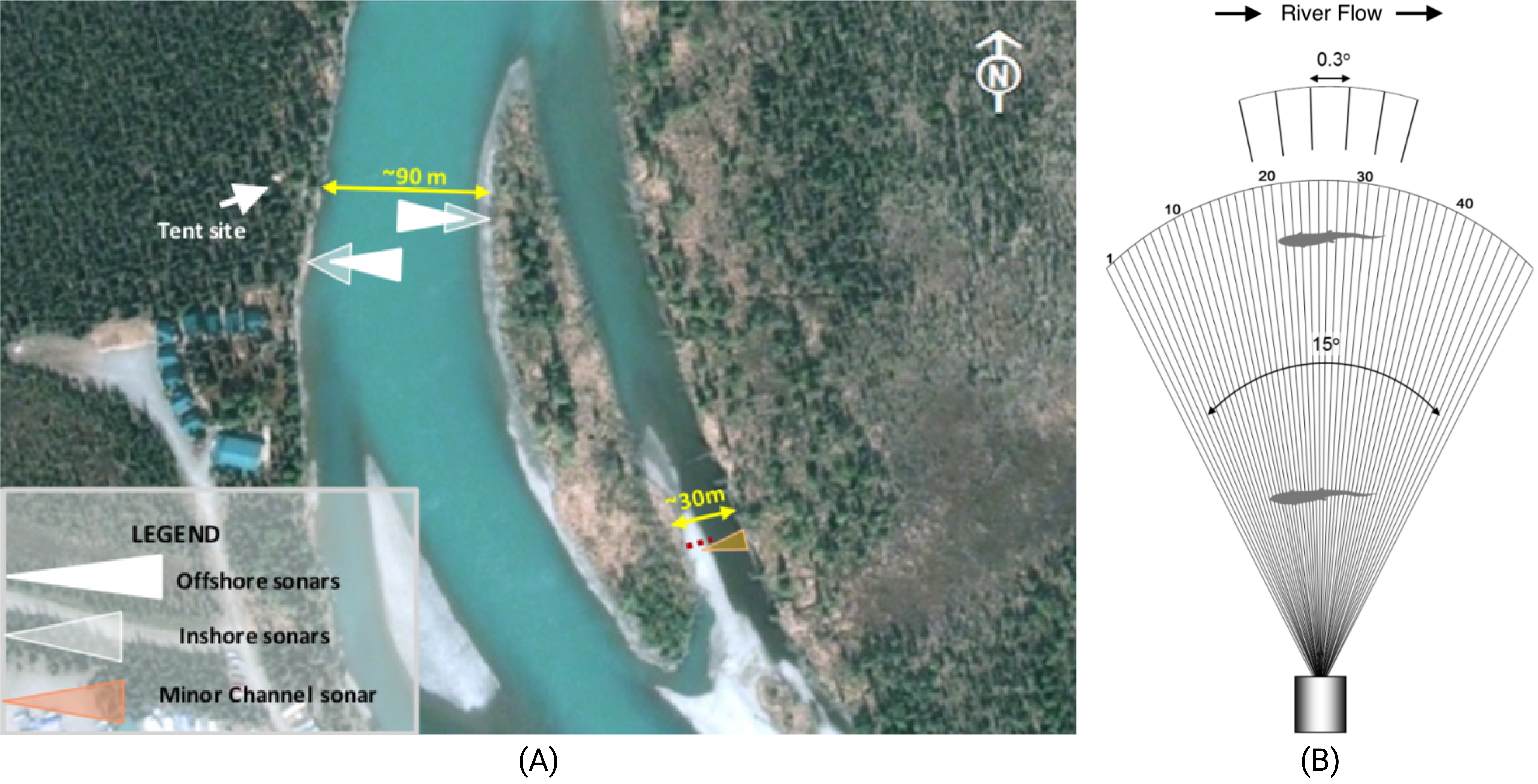}
    \caption{\textbf{(A) Sonar camera configuration on the Kenai River.} The left and right sides of the mainstem, known as Kenai Left Bank (KL) and Kenai Right Bank (KR), respectively, contain two cameras apiece: one near-range and one far-range. KL and KR capture nearby, but not overlapping, areas of the river. Another camera is deployed in the ``minor channel.'' \textbf{(B) A depiction of multi-beam sonar.} Each $3^{\circ}$ segment corresponds to a single beam of sonar. Fish at closer range will be higher resolution since beams spread at distance to cover more area. Both images sourced from \cite{kenaiplan}.}
    \label{fig:sonar_config}
\end{figure}

\begin{figure}[t]
    \centering
    \includegraphics[width=\textwidth]{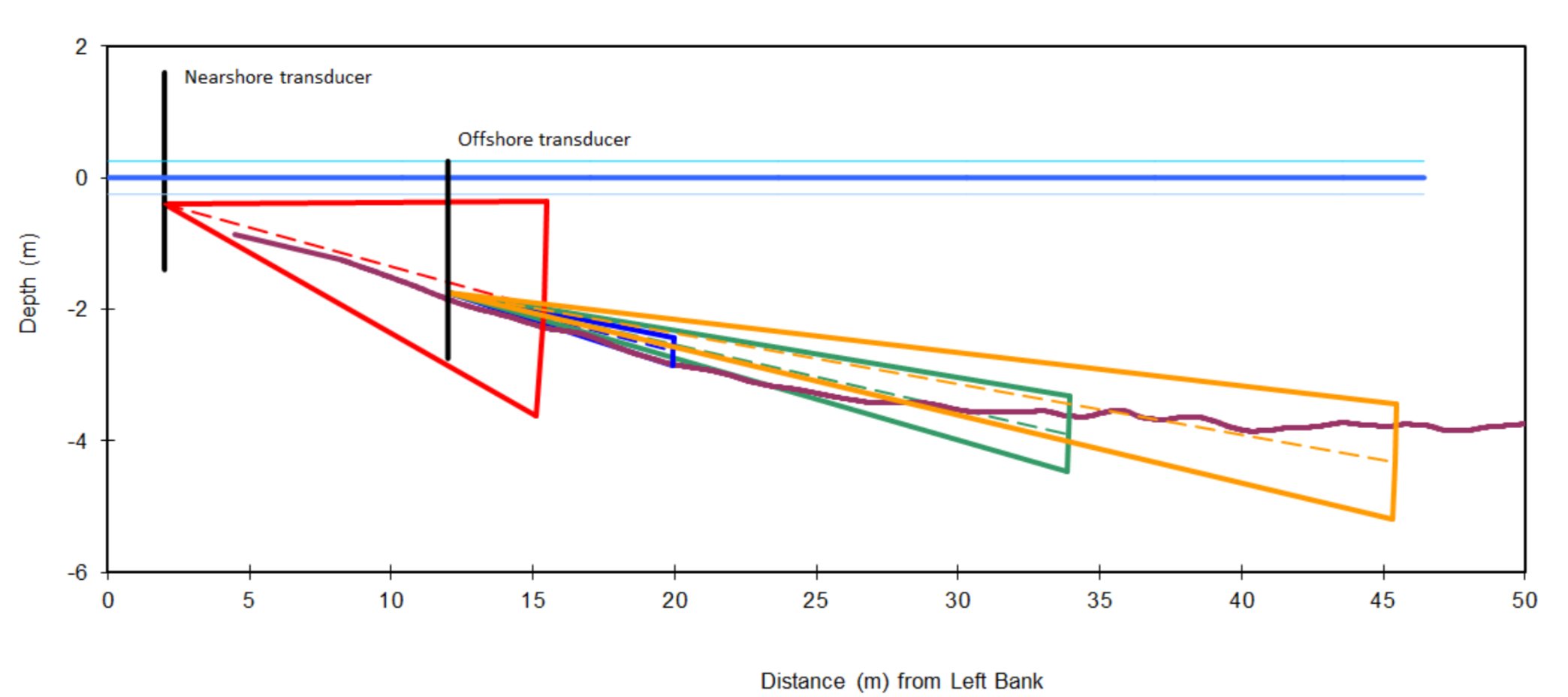}
    \caption{\textbf{An underwater depiction of the sonar camera configuration at Kenai Left Bank (KL).} The contour of the river bottom is shown, along with one triangle depicting the area captured by the near-range camera (leftmost triangle), and three triangles depicting the areas captured by the three strata of the far-range camera. The pitch of each camera/stratum affects the area where fish can be detected as well as how much bottom texture is captured. Source: \cite{kenaiplan}.}
    \label{fig:underwater}
\end{figure}

\begin{figure}[t]
    \centering
    \includegraphics[width=\textwidth]{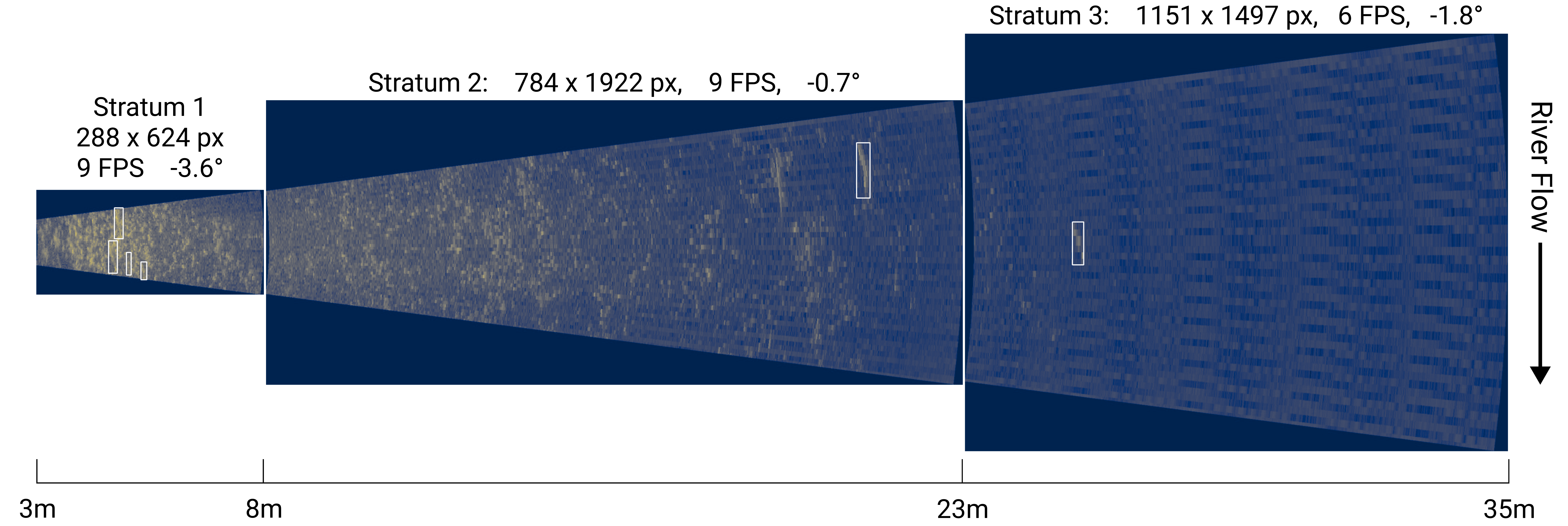}
    \caption{\textbf{A detailed depiction of example frames from each stratum of the far-range camera at Kenai Right Bank (KR).} Ground truth fish detections are shown in white. Each stratum captures varying amounts of bottom information due to difference in pitch, shown in degrees. The image dimensions scale according to the real-world scattering of the beams. At far range (Stratum 3), there are noticable artifacts from the beam scattering, and framerate has been reduced in order to increase sampling frequency, a common tradeoff.}
    \label{fig:rightfar}
\end{figure}

In this section we provide additional background information regarding the imaging sonar format and causes of the challenges enumerated in Fig.~\ref{fig:example} of the main paper. 

The \datasetfullname (\datasetname) consists of video clips sourced from adaptive resolution imaging sonar (ARIS) hardware manufactured by Sound Metrics Corporation. Imaging sonars use an array of sound beams to produce underwater images (see Fig. \ref{fig:sonar_config}B), and have been used to monitor migrating salmon populations in rivers since 2002 \cite{kenaiplan,belcher2002dual}. Because sound travels much further than light in water \cite{simmonds2008fisheries}, sonar can be used to make observations at longer ranges compared to photographic systems and is more robust to turbid conditions \cite{moursund2003fisheries,simmonds2008fisheries}. It can also be used at night, when salmon often travel to minimize predation risk \cite{farrell2011encyclopedia}. The latest generations of imaging sonar have the additional benefit of encoding information about real-world distances, i.e. each pixel in the resulting image represents a defined distance in meters. This can be used to directly measure the length of the observed fish, an important attribute for fisheries management programs in differentiating between species, age groups, and assessing overall population makeup \cite{helminen2020length}.

The ARIS hardware is typically positioned perpendicular to the river flow, thus the fish are observed as moving left-to-right or right-to-left once the sonar data is transformed into video. There are a number of factors which impact the visual characteristics and contribute to a large variance in the images produced at different deployments.

\subsubsection{Environmental factors} include the presence of sediment, floating or static debris, textured river bottom, and plant material. These affect visibility and cause occlusion of target fish. These factors change day-to-day due to weather conditions, as well as on longer time scales due to seasonal variations and anthropogenic impact \cite{thompson2019anthropogenic}. The shape and size of the river, which change throughout the season as water levels rise and fall, have an impact on image quality as well. Smaller river channels can result in increased echo, and variations in the shape of the river bottom can make it difficult for a single camera to effectively observe a broad area, e.g. to cover a steep bank while maintaining visibility further out into the middle of the river.

\subsubsection{Hardware settings} include sonar frequency and camera orientation, which are configured based on the monitoring objectives and river characteristics of each deployment. The operating frequency of a system, typically ranging between 0.7 MHz and 1.8 MHz, determines the image resolution of the recorded video as well as its range capabilities \cite{belcher2002dual,helminen2020length}. There are trade-offs in setting this frequency, since higher frequency sound waves allow for higher resolution imagery, but reduce the range capabilities due to increased sound absorption \cite{simmonds2008fisheries}. Some hardware systems also allow users to choose the number of sonar beams used. More beams help improve resolution at the cost of reducing the observed range as well as the number of frames captured per second \cite{helminen2020length}.

As each beam of sound travels away from the device its observed area increases, decreasing the resolution of longer-range observations. See Fig. \ref{fig:sonar_config}B. In practice, cameras are often cycled through two or more range settings known as \textit{strata} to allow a single camera to sample both near-range and far-range data at specified intervals. See Fig. \ref{fig:rightfar}. For example, cameras on the Kenai River cycle through 2 or 3 strata each hour, such that each camera records 20 minutes of data from a near-range setting, 20 minutes of data from a mid-range setting, and (optionally) 20 minutes of data from a long-range setting \cite{kenaiplan}.

Image quality is also impacted by the camera’s position and orientation in relation to the river floor. A camera which is closer to or pitched toward the bottom will allow for observing fish which swim in deeper waters, however more background information will be picked up which can lead to severe occlusion issues. See Fig. \ref{fig:underwater}.

\subsubsection{Acoustic properties} can cause confounding visual phenomena which interfere with accurate fish observation and measurement. Speckle noise, caused by returning wave interference within the sonar transducer \cite{jaybhay2015study,karabchevsky2011fpga,grabek2019speckle}, makes for grainy imagery which can make fish detection difficult. Acoustic shadows can occlude fish. Sound waves can bounce back and forth between the river bottom and the water surface before returning to the transducer, resulting in multiple shifting bottom images \cite{faulkner2015feasibility}. This type of echo can also create multiple images of individual fish offset from each other, known as ``ghost fish" \cite{faulkner2015feasibility}. The presence and extremity of these ghost images varies based on river shape, camera orientation, and water level.

\section{Dataset}

\begin{figure}[t]
    \centering
    \includegraphics[width=\textwidth]{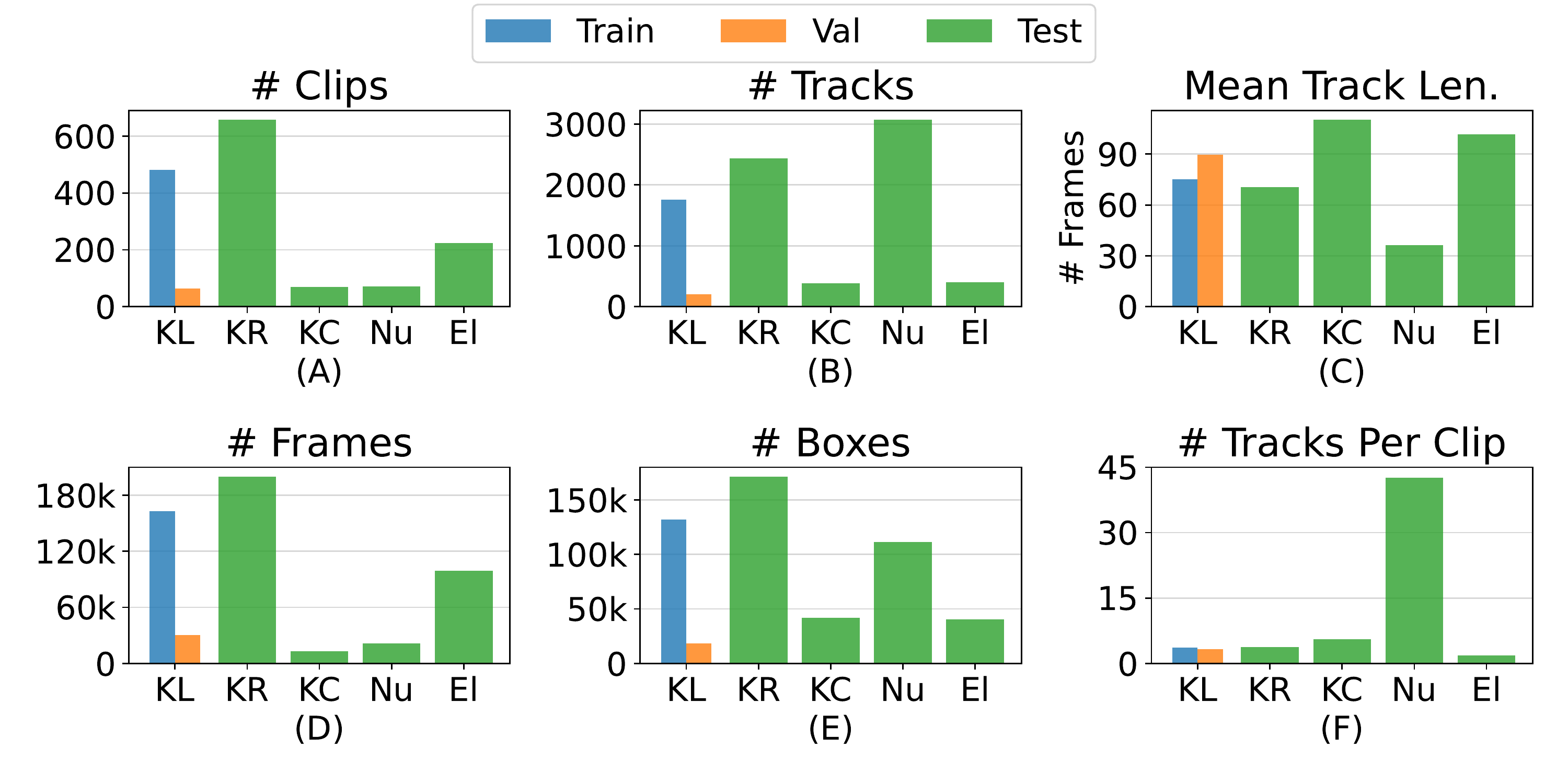}
    \caption{\textbf{Dataset statistics across locations and data split.} \textbf{(A)}~Number of video clips. \textbf{(B)}~Number of annotated tracks. \textbf{(C)}~Average track length in frames. \textbf{(D)}~Total number of video frames. \textbf{(E)}~Number of annotated bounding boxes. \textbf{(F)}~Average number of tracks per video clip. \textbf{KL}:~Kenai Left Bank. \textbf{KR}:~Kenai Right Bank. \textbf{KC}:~Kenai Channel. \textbf{Nu}:~Nushagak. \textbf{El}:~Elwha.}
    \label{fig:loc_stats}
\end{figure}

\begin{figure}[t]
    \centering
    \includegraphics[width=\textwidth]{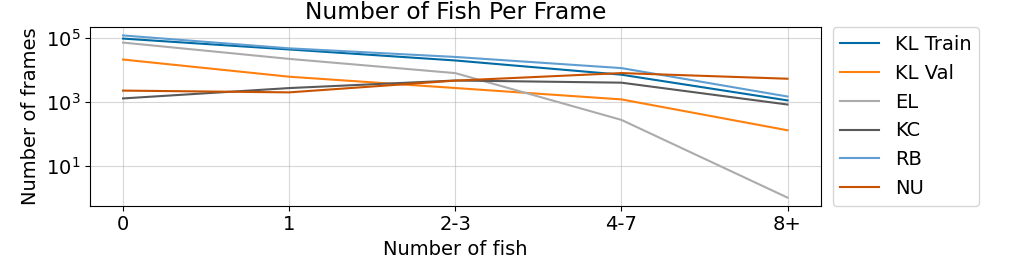}
    \caption{\textbf{Histogram of fish per frame (including empty frames) at each location.} Note the large number of fish at the Nushagak location (NU), posing a tough generalization challenge for methods.}
    \label{fig:loc_stats_rebuttal}
\end{figure}

In this section we provide additional details about the data source locations in \datasetname. The distribution of video clips and annotations can be seen in Fig.~\ref{fig:loc_stats}, and generalization challenges from clip density can be seen in Fig.~\ref{fig:loc_stats_rebuttal}.

\subsubsection{Kenai River (AK)} We received data from 5 different cameras deployed at 3 locations on the Kenai River at river mile (RM) 13.7 between May 26th and August 17th 2018. The left and right sides of the river mainstem contain one near-range camera and one far-range camera apiece. In our dataset we refer to data from the left and right sides of the main river channel as KL (Kenai Left Bank) and KR (Kenai Right Bank) respectively. There is an additional smaller tributary at RM 13.7 known as the ``minor channel'', where a fifth camera is placed, the data from which we refer to as KC (Kenai Channel).

Of the data we received, the Kenai River had the most useful manual annotations, specifying a frame number, location in polar coordinates, and size measurement for each observed fish. To create the dataset for this location we extracted 200-frame video clips centered around randomly-selected manually marked frames. This duration was chosen after initial visual inspection to approximate the time it took for fish to enter and exit the field of view. If any of these clips overlapped, we merged them into one longer clip. In total we extracted 1233 clips from the Kenai River containing 4300 fish.
 
\subsubsection{Nushagak River (AK)} We received data from 2 cameras deployed on the Nushagak River approximately 57 river kilometers (RKM) from the town of Dillingham between June 7th and September 1st 2018. We received hourly counts rather than timestamps of fish observations, so manual inspection was required to find clips containing fish. In one of the cameras, fish were very sparse and it was prohibitively time-consuming to do. We focused instead on the second camera, in which fish were very abundant (see Fig.~\ref{fig:loc_stats}F). We sampled 72 300-frame clips containing 3070 fish. 
%The Nushagak River annotations did not contain size measurements, thus this data does not appear in our length measurement benchmarks.
    
\subsubsection{Elwha River (WA)} We received data from a single camera deployed at RKM 1 of the Elwha River between July 9th and September 18th 2018. Timestamps and length measurements for all fish were also provided. We used the same protocol as the Kenai, randomly sampling 200-frame clips and merging any that overlapped. In total we sampled 262 clips containing 884 fish. Compared to the other locations, the Elwha data is much more sparse due to a nascent recovery of its salmon populations (see Fig.~\ref{fig:loc_stats}B). This river is currently rebounding from a 90\% reduction in salmon populations as a result of damming in the early 1900s \cite{pess2008biological}, and after the largest dam removal project in history is beginning to see the return of several species \cite{fraik2021impacts}.

\section{Additional Experiments}

\subsection{Detector Architecture Search}

\setlength{\tabcolsep}{4pt}
\begin{table}[t]
\begin{center}
\caption{\textbf{Object detector comparison on the \datasetname validation set.} These detectors were trained and evaluated on raw sonar frames. We chose YOLOv5m for our Baseline methods due to its superior performance}
\label{table:detectors}
\begin{tabular}{l|l}
\hline\noalign{\smallskip}
Detector & Validation AP50 \\
\noalign{\smallskip}
\hline\hline
\noalign{\smallskip}
Faster R-CNN + Resnet101 & 65.9 \\
ScaledYOLOv4 CSP & 65.3 \\
YOLOv5m & \textbf{66.4} \\
\hline
\end{tabular}
\end{center}
\end{table}
\setlength{\tabcolsep}{1.4pt}

We benchmarked three state-of-the-art object detection architectures: Faster R-CNN \cite{ren2015faster} with a Resnet-101 \cite{he2016deep} backbone; ScaledYOLOv4 \cite{wang2021scaled}; and YOLOv5 \cite{glenn_jocher_2022_6222936}. We trained on the \datasetname training set and evaluated on the validation set. For these experiments we used the raw verison of the sonar frames (i.e. not the novel input format used by the Baseline++ method). We selected our final architecture based on validation set AP50. See Table \ref{table:detectors}.

\subsubsection{Faster R-CNN Training Settings} 
We fine-tuned a Faster R-CNN model pretrained on COCO using the default training settings from Detectron2 v0.6 \cite{wu2019detectron2} with the following modifications: we trained with a batch size of 8 and a learning rate of 0.0025 on two NVIDIA RTX A5000 GPUs for 18 epochs, reducing the learning rate by a factor of 10 at epochs 12 and 16, and selected the best model checkpoint based on validation AP50. 

\subsubsection{ScaledYOLOv4 Training Settings} We fine-tuned a ScaledYOLOv4 CSP model pretrained on COCO using the default training settings from the official implementation with the following modifications: we resized all inputs to 896px on their longest side and selected the best model checkpoint based on validation AP50. We used a batch size of 32 and trained on two NVIDIA RTX A5000 GPUs.

\subsubsection{YOLOv5 Training Settings} We fine-tuned a YOLOv5m model pretrained on COCO using the default training settings from v6.0 release with the following modifications:  we resized all inputs to 896px on their longest side, trained the detector for 150 epochs, and selected the best model checkpoint based on validation AP50. We used a batch size of 64 and trained on two NVIDIA RTX A5000 GPUs.

\subsection{Appearance-based Re-ID}
\label{sec-reid}

\setlength{\tabcolsep}{4pt}
\begin{table}
\begin{center}
\caption{\textbf{Resnet-50 + Triplet Loss re-identification performance on \datasetname validation set compared to several common re-identification benchmarks.} The same network performs very poorly on \datasetname, verifying that appearance features are not a strong signal for association during tracking. * indicates results reported from \cite{hermans2017defense}, which used test-time augmentation and no training-time augmentation. All other training and evaluation settings match ours}
\label{table:reid}
\begin{tabular}{l|l}
\hline\noalign{\smallskip}
Dataset & mAP \\
\noalign{\smallskip}
\hline\hline
\noalign{\smallskip}
CUHK03 \cite{li2014deepreid} & 80.7 \\
Market1501 \cite{zheng2015scalable} & 67.9 \\
MARS* \cite{zheng2016mars} & 67.7 \\
\midrule
\datasetname (Val) & 19.2  \\
\hline
\end{tabular}
\end{center}
\end{table}
\setlength{\tabcolsep}{1.4pt}

%p@1, map, map@r
%Resnet50 + Triplet Loss & 34.7 & 19.2 & 13.8\\

Our baseline tracker \cite{bewley2016simple} uses a motion model and a simple IoU metric to perform association. It is also common to incorporate appearance information into the association costs \cite{ciaparrone2020deep,luo2021multiple,zheng2019joint}. However, due to the lack of differentiating features between individual fish in \datasetname, we did not expect appearance-based re-identification methods to work well for association. 

We verified this by implementing a popular visual re-identification network inspired by \cite{hermans2017defense}, based on a Resnet-50 and Triplet Loss. We show in Table \ref{table:reid} that it is indeed much less effective on \datasetname than on standard re-identification datasets. We trained the model by cropping out ground truth detections from our training set. Crops from the same ground-truth track were considered positive re-identification matches. We trained using the Adam optimizer \cite{kingma2014adam} with a learning rate of 0.005 and batch size of 128 for 50 epochs. We used an output embedding size of 128 and the default data augmentations as described in \cite{musgrave2020metric,openreid}. Results for other datasets are reported from \cite{openreid} and \cite{hermans2017defense}.

\subsection{Baseline Results}

\begin{figure}
    \centering
    \includegraphics[width=\textwidth]{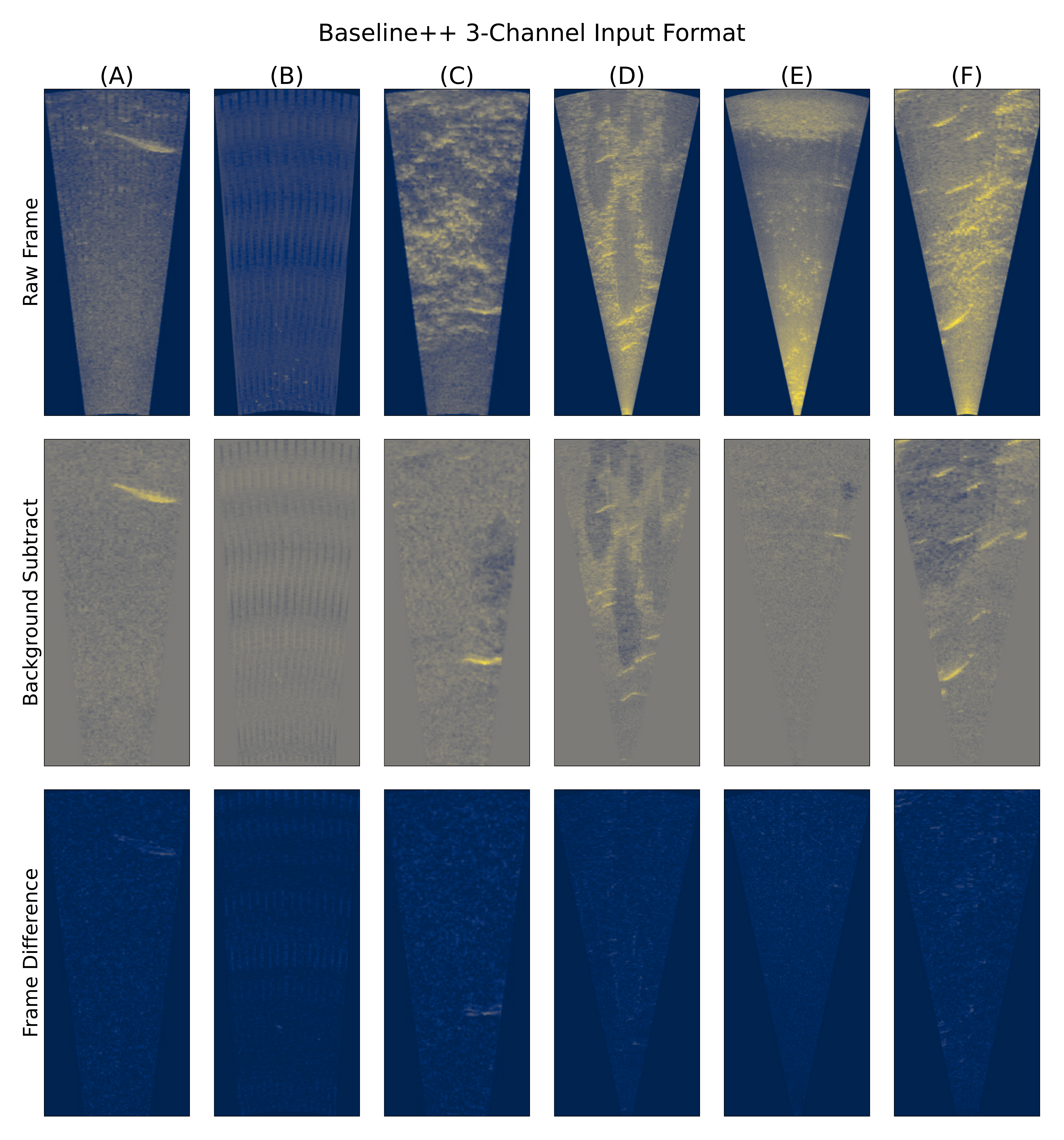}
    \caption{\textbf{Example frames using the enhanced input format in Baseline++ method.} Our Baseline++ method appends a background-subtracted channel and a frame-differenced channel to the raw input frame (see Sec. 5.3 of the main paper). Shown here are the same example frames from Fig.\ref{fig:example}A--F of the main paper.}
    \label{fig:3channel}
\end{figure}

% Both baselines in one table %
\setlength{\tabcolsep}{4pt}
\begin{table}[t]
\begin{center}
\caption{\textbf{Baseline and Baseline++ results on \datasetname}
}
\label{table:overall_results}

%% Original - without HOTA %%
% \begin{tabular}{l|l|l|l|l||l|l|l|l}
% \hline\noalign{\smallskip}
% \multirow{2}{*}{Location} & \multicolumn{4}{c||}{Baseline} & \multicolumn{4}{c}{Baseline++} \\
% & AP$\uparrow$ & MOTA$\uparrow$ & IDF1$\uparrow$ & nMAE$\downarrow$ & AP$\uparrow$ & MOTA$\uparrow$ & IDF1$\uparrow$ & nMAE$\downarrow$ \\
% \noalign{\smallskip}
% \hline\hline
% \noalign{\smallskip}
% KL (Val) &66.4 &44.9 &66.7 &4.9\% &68.0 &47.8 &68.5 &3.3\% \\
% \midrule
% KR & 57.7 &-28.5 &45.4 &11.8\% &87.1 &69.8 &82.3 &3.7\% \\
% KC & 32.0 &-60.8 &35.6 &53.0\% &65.1 &44.3 &65.7 &12.8\% \\
% NU & 70.6 &30.2 &60.8 &14.0\% &85.5 &56.2 &75.1 &8.6\% \\
% EL & 39.9 &-376.7 &18.8 &32.3\% &74.7 &-54.5 &47.1 &21.3\% \\
% \hline
% \end{tabular}

%% Adding HOTA -- To eliminate overflow: removed center double divider; removed up and down arrows; changed "Location" to "Loc" and "KL (Val)" to "KL \\ Val" %%
\begin{tabular}{r|r|r|r|r|r|r|r|r|r|r}
\hline\noalign{\smallskip}
\multirow{2}{*}{Loc} & \multicolumn{5}{c|}{Baseline} & \multicolumn{5}{c}{Baseline++} \\
& \multicolumn{1}{c|}{AP} & MOTA & IDF1 & HOTA & nMAE & \multicolumn{1}{c|}{AP} & MOTA & IDF1 & HOTA & nMAE \\
\noalign{\smallskip}
\hline\hline
\noalign{\smallskip}
\multicolumn{1}{l|}{KL} & \multirow{2}{*}{66.4} & \multirow{2}{*}{44.9} & \multirow{2}{*}{66.7} & \multirow{2}{*}{49.2} & \multirow{2}{*}{4.9\%} & \multirow{2}{*}{68.0} & \multirow{2}{*}{47.8} & \multirow{2}{*}{68.5} & \multirow{2}{*}{51.2} & \multirow{2}{*}{3.3\%} \\
\textit{Val} & & & & & & & & & & \\
\midrule
\multicolumn{1}{l|}{KR} & 57.7 &-28.5 &45.4 &33.5 &11.8\% &87.1 &69.8 &82.3 &60.3 &3.7\% \\
\multicolumn{1}{l|}{KC} & 32.0 &-60.8 &35.6 &30.9 &53.0\% &65.1 &44.3 &65.7 &49.0 &12.8\% \\
\multicolumn{1}{l|}{NU} & 70.6 &30.2 &60.8 &44.4 &14.0\% &85.5 &56.2 &75.1 &54.4 &8.6\% \\
\multicolumn{1}{l|}{EL} & 39.9 &-376.7 &18.8 &21.3 &32.3\% &74.7 &-54.5 &47.1 &38.6 &21.3\% \\
\hline
\end{tabular}

\end{center}
\end{table}
\setlength{\tabcolsep}{1.4pt}

\begin{figure}
    \centering
    \includegraphics[width=\textwidth]{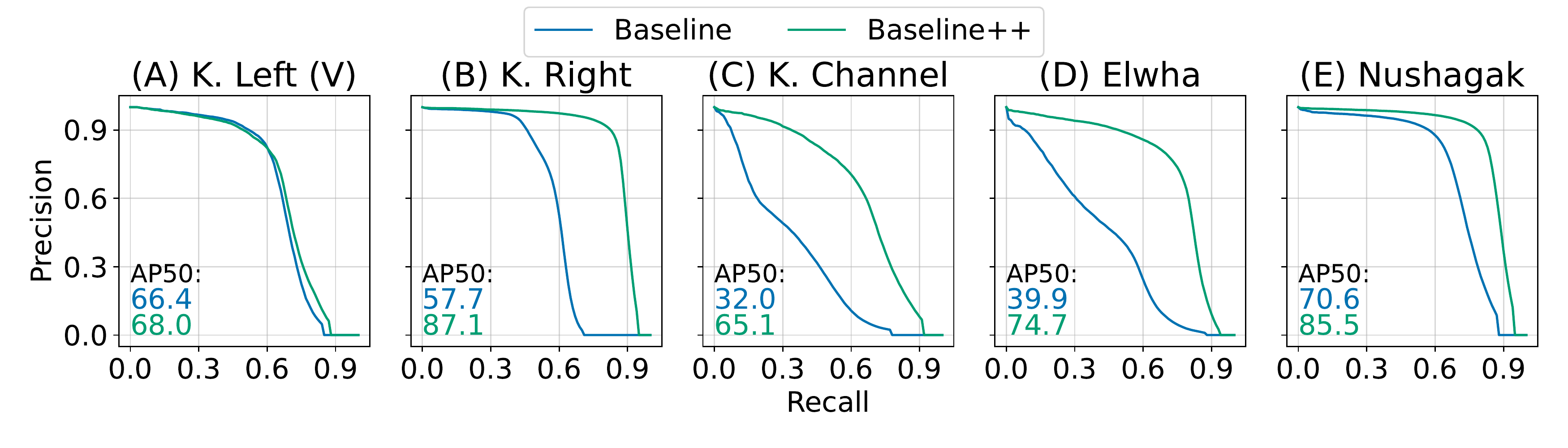}
    \caption{\textbf{Object detection precision/recall curves.} Performance comparison between Baseline and Baseline++ method at each location in \datasetname.}
    \label{fig:detectors}
\end{figure}

Fig.~\ref{fig:3channel} shows example frames illustrating the 3-channel input format used by our Baseline++ method. We append two additional channels to the raw input frame: a background-subtracted channel obtained by subtracting the clip-wise average frame, and a frame-differenced channel obtained by taking the difference of the background-subtracted versions of the current and previous frame.

Tab.~\ref{table:overall_results} shows the full results for our Baseline and Baseline++ methods across all metrics. We include additional visualizations of our Baseline and Baseline++ object detection performance in Figure \ref{fig:detectors}. While the improvements for the Baseline++ method are marginal at the training/validation location (Kenai Left Bank, Fig. \ref{fig:detectors}A), improvements are significant at the out-of-sample test locations, as indicated by the large increase in area under the PR curve in Fig. \ref{fig:detectors}B--E.

\clearpage
% ---- Bibliography ----
%
% BibTeX users should specify bibliography style 'splncs04'.
% References will then be sorted and formatted in the correct style.
%
% \bibliographystyle{splncs04}
% \bibliography{bib/behavior,bib/counting,bib/domain_adapt,bib/fgvc,bib/other,bib/reid,bib/sdm,bib/tracking}

\end{document}